\documentclass[twocolumn,final]{svjour3} 
\smartqed  

\usepackage[export]{adjustbox}
\usepackage{blkarray, bigstrut}
\usepackage{xparse}

\usepackage{multirow} 
\usepackage{algorithmicx}
\usepackage{algpseudocode}
\usepackage{algorithm}
\usepackage{amsmath}
\usepackage{rotating}
\usepackage{tabularx}

\usepackage{mathtools}
\usepackage{amsmath}
\usepackage{stfloats}

\usepackage{kantlipsum} 
\usepackage{commath}
\allowdisplaybreaks
\usepackage{mathtools}  
\usepackage{verbatim}
\usepackage{xr-hyper} 
\usepackage{enumitem}
\usepackage{multirow}
\usepackage{hhline}
\usepackage[table,xcdraw]{xcolor}
 \usepackage{graphicx,booktabs}
 \usepackage{longtable}
 \usepackage{subcaption}

\usepackage[hyphens]{url}
\usepackage{hyperref}
\hypersetup{breaklinks=true}
\usepackage{epstopdf}
\usepackage{wrapfig}
\usepackage{latexsym}
\usepackage{amssymb}
\usepackage{amsfonts}
\usepackage{amsmath} 
\usepackage{graphicx}
\usepackage{latexsym}
\usepackage{booktabs}
\usepackage[style=base]{caption}
\usepackage{subcaption} 
\usepackage{breqn}
\newtheorem{thm}{Definition}

\definecolor{white}{rgb}{0.88,1,1}
\algnewcommand\algorithmicinput{\textbf{INPUT:}}
\algnewcommand\INPUT{\item[\algorithmicinput]}
\algnewcommand\algorithmicoutput{\textbf{OUTPUT:}}
\algnewcommand\OUTPUT{\item[\algorithmicoutput]}
\algdef{SE}[DOWHILE]{Do}{doWhile}{\algorithmicdo}[1]{\algorithmicwhile\ #1}

\makeatletter
\newcounter{algorithmbis}
\renewcommand{\thealgorithmbis}{\arabic{algorithmbis}}
\def\algorithmbis{\@ifnextchar[{\@algorithmbisa}{\@algorithmbisb}}
\def\@algorithmbisa[#1]{%
  \refstepcounter{algorithmbis}
  \trivlist
  \leftmargin\z@
  \itemindent\z@
  \labelsep\z@
  \item[\parbox{0.49\textwidth}{%
    \hrule
    \noindent\strut\textbf{Algorithm \thealgorithmbis} #1
    \hrule
  }]\hfil\vskip+0em%
}
\def\@algorithmbisb{\@algorithmbisa[]}

\makeatother

\usepackage[table]{xcolor}

\usepackage{tikz}
\usetikzlibrary{fit} 

\usepackage{color}

\colorlet{BLUE}{blue}
\usepackage{colortbl}
\definecolor{white}{RGB}{155, 227, 247}

\newif\ifcommentson

\newif\ifextended
\newif\ifshortver

\shortvertrue

\newcommand{\optional}[1]{\ignorespaces}

\newif\ifrevisionactive
\newif\ifshowdeleted
\revisionactivetrue
\allowdisplaybreaks

\definecolor{maroon}{cmyk}{0,0.87,0.68,0.32}

\begin{document}

\title{On Defending Against Label Flipping Attacks on Malware Detection Systems}

\titlerunning{On Defending Against Label Flipping Attacks on Malware Detection Systems}


	\author{Rahim~Taheri\and Reza~Javidan\and Mohammad~Shojafar\and Zahra~Pooranian\and Ali~Miri\and Mauro~Conti}

\institute
    {
    R. Taheri and R. Javidan \at Department of Computer Engineering and Information Technology, Shiraz University of Technology, Shiraz, Iran\\\email{\{r.taheri, javidan\}@sutech.ac.ir}
	\and
	M. Shojafar, Z. Pooranian and M. Conti \at SPRITZ, Department of Mathematics,\\ University of Padua, Padua, Italy\\
		\email{mohammad.shojafar@unipd.it; m.shojafar@ieee.org}\\\email{\{zahra, conti\}@math.unipd.it}
	\and
	A. Miri \at Department of Computer Science,\\ Ryerson University, Toronto, Canada\\
		\email{ali.miri@ryerson.ca}
}
\date{Received: 23 July 2019 / Revised: 22 February 2020/ Accepted: 04 March 2020}
\maketitle

\begin{abstract}

Label manipulation attacks are a subclass of data poisoning attacks in adversarial machine learning used against different applications, such as malware detection. These types of attacks represent a serious threat to detection systems in environments having high noise rate or uncertainty, such as complex networks and Internet of Thing (IoT). Recent work in the literature has suggested using the $K$-Nearest Neighboring (KNN) algorithm to defend against such attacks. However,  \textcolor{black}{such an approach can suffer from low to miss-Classification rate Accuracy}.
In this paper, we design an architecture to tackle the Android malware detection problem in IoT systems. We develop an attack mechanism based on Silhouette clustering method, modified for mobile Android platforms. We proposed two Convolutional Neural Network (CNN)-type deep learning algorithms against this \emph{Silhouette Clustering-based Label Flipping Attack (SCLFA)}. We show the effectiveness of these two defense algorithms - \emph{Label-based Semi-supervised Defense (LSD)} and \emph{clustering-based Semi-supervised Defense (CSD)} - in correcting labels being attacked. We evaluate the performance of the proposed algorithms by varying the various machine learning parameters on three Android datasets: Drebin, Contagio, and Genome and three types of features: API, Intent, and Permission. Our evaluation shows that using random forest feature selection and varying ratios of features can result in an improvement of up to 19\% \textcolor{black}{Accuracy} when compared with the state-of-the-art method in the literature.



\end{abstract}
\section{Introduction}\label{sez:1}
Machine learning (ML) algorithms have the ability to accurately predict patterns in data. However, some of the data can come from uncertain and untrustworthy sources. Attackers can exploit this vulnerability as part of what is known as \emph{Adversarial Machine Learning (AML)} attacks. \emph{Poisoning attacks} or \emph{data poisoning attacks} are a subclass of AML attacks, in which attackers inject malicious data into the training set in order to compromise the learning process, and effect the algorithm performance in a targeted manner. 
\textit{Label flipping} attacks are a special type of data poisoning, in which the attacker can control labels assigned to a fraction of training points. Label flipping attacks can significantly diminishes the performance of the system, even if the attacker's capabilities are otherwise limited. 
Recent work in AML looks into  effectiveness of poisoning attacks in degrading performance of popular classification algorithms, such as support vector machines (SVM)~\cite{zhou2012adversarial}, embedded features selection methods~\cite{xiao2015feature,zhang2016adversarial}, neural networks~\cite{ganin2016domain}, and deep learning systems~\cite{papernot2016distillation}. Most attacks in the literature assume attackers can manipulate both  features and labels associated with the poisoning data. However, sometimes the attacker's capabilities are limited to manipulating labels, and he is only able to flip the labels to fool the ML classifier. These types of attacks are known  as \textit{flipping attacks}.
Deep neural networks (DNNs) have gained significant success in classifying well labeled data. However, label flip type poisoning attacks can reduce the \textcolor{black}{Accuracy} of these algorithms~\cite{yang2017generative}. Therefore, there is a need for alternative methods for 
training DNNs that takes label flipping attacks into account. Such methods should be able to identify and correct mislabeled samples or re-weight the data terms in the loss function according to the extracted label. 

There are a number of work in the literature focused on identifying and dealing with poisoning attacks. For example, an algorithmic method evaluates the impact of each training sample on the performance of learning algorithms~\cite{bhagoji2017dimensionality}. Although this method is effective in some cases, it cannot be generalized to the large dataset. Among other defensive mechanisms, the \textit{outlier detection} is used to identify and remove suspicious samples. But, this method has a limited performance (i.e., \textcolor{black}{Accuracy}) against label flipping attacks~\cite{paudice2018label}. Another category of related works mainly focus on learning strategies that can be applied on flip labels. Such solutions are divided into \textit{two} categories. In the first group, it can directly learn flipped labels, whilst in the second group, it can focus on an extra set of clean data. In the first case, the label flipping module is used to identify correctly labeled data~\cite{natarajan2013learning,xiao2015support}, and to modify the changes on the labels to reset the data terms in the loss function.
 Performance of this technique is significantly impacted by its label cleaning \textcolor{black}{Precision} and its rate of flip sample estimation. In the second group of methods, an additional set of clean data is used to guide the learning agent through flipped data~\cite{ren2018learning}. Despite promising results, both groups of methods have a common default. They try to fix the flipped labels, or they re-weight the terms for data points. This default will inevitably cause errors for some data points.
 
Motivated by these considerations, in this paper, we consider the \textit{binary classification} for sampling and analysis of Android malware. We only assume the weakest capability for the attacker. That is, we assume that the attacker has no perfect knowledge about the learning algorithm, the loss function optimized by the system, or the initial the training data and a set of features used by the learning algorithm. We show that having the system identifying and retraining the wrong label, and using our proposed Semi-Supervised (SS) approach to training will result in better results. To this end, we suggest a solution that covers the existing data points that are mislabeled and improves the \textcolor{black}{Accuracy} of the classification algorithm. To do so, we present an architecture for learning flipped data. Then, we identify a small part of the mislabeled training set, whose labels are likely to be correct, and the flipped labels associated with other data are ignored. Afterward, we train a deep neural network in a SS manner based on selected data.

\subsection{Contributions}        
In this context, several natural questions are arising, such as: How can we define attack based on label flipping algorithm which can fool the classifier? Is it possible to design an enhanced ML model to improve system security by presenting some secure algorithms against a given label flipping attack? How can we tune and test the countermeasure solutions to deal with label flipping attack? The answer to these queries is the goal of this paper. \textcolor{black}{More in detail, the goal of the paper summarizes as follows: First, we rank the data points within each class and then hold the label for the points that have higher rankings. If no clean set is available, the ranking is based on the multi-way classification neural network, which is trained from the original training dataset. In fact, a binary classifier is learned that, while clean labels are available, separates data containing clean labels and flipped labels. Second, we apply a temporary ensemble for semi-supervised deep neural network training. Hence, our original contributions are as follows:}

     \begin{itemize}[leftmargin=*]
        \item \textcolor{black}{ We present an architecture for learning flipped data which reflects our main focus in the malware detection system.}
       \item  \textcolor{black}{We propose a label flipping poisoning technique to attack the Android malware detection based on deep learning: where an algorithm is proposed for crafting efficient prototypes so that the attacker can deceive the classification algorithm. In this technique, we use Silhouette clustering to find an appropriate sample to flip its label.}
        \item \textcolor{black}{We introduce a DL-based semi-supervised approach against label flipping attacks in the malware detection system called LSD, which uses label propagation and label spreading algorithms along with CNNs to predict the correct value of labels for the training set.}
        \item \textcolor{black}{We implement a countermeasure method based on clustering algorithms as a defense mechanism. It is a DL-based semi-supervised approach against label flipping attacks in the malware detection system that improves the detection Accuracy of the compromised classifier. In this approach we use \textit{four} clustering metrics and validation data to re-labeled poisoned labels.}
       \item \textcolor{black}{We conduct our experiments on two scenarios on three real Android datasets using three feature types compared to the cutting-edge method and deeply analyze the trade-offs that emerge. The source code of the paper is available in Github~\cite{Teheri2020NCAA}.}
    \end{itemize}
    
To best of our knowledge, none of the previous works in literature has conducted a similar analysis. The closet paper to our method is KNN-based Semi-Supervised Defense (KSSD)~\cite{paudice2018label}, in which the authors have entailed KNN strategy to relabel samples by considering the distance between them. However, the work in~\cite{paudice2018label} is tailored to the relabeling of samples, they are unable to specify some similar samples that may be malware and benign and may mislabel the features of benign sample due to low distance of samples. Unlike the~\cite{paudice2018label}, in this paper we explicitly tackle the poisoning samples located far from the decision boundary and relabel them. Also the defense method presented in~\cite{paudice2018label} is unable to distinguish overlapping areas of two classes and cannot correctly label the poisoning samples located there while our defense methods imposes the model to tackle such data points and relabeling them. 

 \subsection{Organization of the paper}
  \textcolor{black}{We organize the rest of the paper as follows. Section~\ref{relatedWork} overview the related works. Section \ref{proposedarchitecture} details the problem definition, the presented architecture, and the related components. Section \ref{proposedsolution} presents our proposed attack model inspired by AML architecture and reports the proposed defense strategies against the raised attack. We evaluate the performance of the algorithms in Section~\ref{resultAnalysis}. In Section~\ref{discussion} we detail the 
  results of the experiment and provide some open discussion regarding our method. Section~\ref{conclusion} presents conclusions and future work. Table ~\ref{tab1} shows the important abbreviations used in this paper.}

\begin{table}[!htpb]
\centering
\caption{\small \textcolor{black}{Important abbreviations used in this paper.}\vspace{-5px}}
\label{tab1}
\scriptsize{
\setlength\tabcolsep{1.5pt} 
\begin{tabular}{|c|
>{\columncolor[HTML]{fcfcf4}}l|}
\hline
\rowcolor{white}
\textbf{Notations}  
&\multicolumn{1}{|c|}{\textbf{Description}}\\ \hline
\cellcolor[HTML]{E8E8AB}\textbf{$AML$}& Adversarial Machine Learning  \\\hline
\cellcolor[HTML]{E8E8AB}\textbf{$SSL$}&semi-supervised learning \\\hline
\cellcolor[HTML]{E8E8AB}\textbf{$LSD$}& Label-based  Semi-supervised  Defense\\\hline
\cellcolor[HTML]{E8E8AB}\textbf{$CSD$}&clustering-based  Semi-supervised  Defense \\\hline
\cellcolor[HTML]{E8E8AB}\textbf{$KSSD$}& KNN-based Semi-Supervised Defense\\\hline
\cellcolor[HTML]{E8E8AB}\textbf{$GAN$}& Generative Adversarial Network\\\hline
\cellcolor[HTML]{E8E8AB}\textbf{$CNN$}& Convolutional Neural NetworK \\\hline
\cellcolor[HTML]{E8E8AB}\textbf{$LP$}& Label  Propagation \\\hline
\cellcolor[HTML]{E8E8AB}\textbf{$LS$}& Label Spreading \\\hline
\cellcolor[HTML]{E8E8AB}\textbf{$RI$}&Rand  Index\\\hline
\cellcolor[HTML]{E8E8AB}\textbf{$MI$}& Mutual  Information \\\hline
\cellcolor[HTML]{E8E8AB}\textbf{$FMI$}& Fowlkes-Mallows Index \\\hline
\end{tabular}}
\end{table}

\section{Related work}\label{relatedWork}
  
  In this section, we classify the related work in the literature into two different defense classes: i) we will cover defense approaches that try to correct labels in Section \ref{sec:2.1}, and ii) defense strategies that ignore poisoned labels and adopt semi-supervised learning methods to protect the model against attacks are then covered in Section \ref{sec:2.2}. Hence, we draw conceptual relationships and delineate the most recent defense strategies applied to tackle the label flipping attack and identify relevant major alternatives for comparison.  

\subsection{Defense algorithms against poisoning attacks}\label{sec:2.1}
The problem of classification with label noise -  mislabeling in class variable - is an active area of research. The paper~\cite{frenay2014comprehensive} gives a comprehensive overview of both the theoretical and applied aspects of this area.\textcolor{black}{Label flipping mechanism is a solution to cover label noise in the classifiers~\cite{bootkrajang2012label}. This method can model the overall label flipping probability. However, it is lack of considering individual specific characteristics in label noise. In~\cite{laishram2016curie}, the authors create a lightweight method called \textit{Curie} to protect SVM Classier against poisoning attacks. The preliminary idea behind this method is to distinguish the suspicious data points and remove them outside the dataset before starting the learning step of the SVM algorithm. In other words, Curie's algorithm flips labels in the training dataset to defend SVM classifiers against poisoning attacks. They cluster the data in the feature space and try to calculate the average distance of each point from the other points in the same cluster with related weight and train model and test in some datasets. They present that their defense method is able to correct 95\% of samples in the training dataset. Additionally, the authors in~\cite{munoz2017towards} describe a poisoning algorithm to solve the bi-level optimization problem based on back-gradient optimization~\cite{maclaurin2015gradient}. The proposed algorithm applies automatic differentiation technique to compute the gradient in the optimization problem. This algorithm using gradient method to resolve the optimization problem which takes several computational time, it can pose challenges in complex networks such as neural networks and deep learning. Thus, they apply a novel technique named back-gradient optimization to allow computing the gradient of interest in a more computationally efficient and stable manner to shape their ML model.} Authors in~\cite{wang2018data} explicitly investigate data poisoning attacks for the semi-online setting, unlike other works which mostly based on the offline setting. The work in~\cite{shafahi2018poison} argues that it is possible to perform targeted attacks on specific testing data without declining the overall performance of classifier along with any control of adversary over the labeling of training data. - The methodology proposed in~\cite{baracaldo2018detecting} is suitable to identify and remove poisonous data in IoT systems. This method, mainly, exploits data provenance to identify manipulated data before the training step to improve the performance of classification. Compared to our method, the defense method presented in~\cite{baracaldo2018detecting} cannot correctly label the poisoning samples while our defense methods imposes the model to tackle such data points and relabeling them. The work in~\cite{bootkrajang2014learning} focus on building an automatic robust multiple kernel-based logistic regression classifier against poisoning attacks without applying any cross-validation. Despite the fact that proposed classifier may improve performance and learning speed; it does suffer from lack of any theoretical guarantees. To address this issue, they extend their method and entail new structure to resist the negative effect of random label noise as well as a wide range of non-random label noises~\cite{bootkrajang2016generalised}.

\subsection{Semi-supervised learning defense algorithms}\label{sec:2.2}

Another active area of research is the one dealing with learning from unlabeled data. The semi-supervised learning approach, along with applying unlabeled data to learn better models is particularly relevant to our work. The semi-supervised approaches include multiview learning like~\cite{dong2018tri}, co-training~\cite{ren2018learning,xia20183d}, graph-based methods like~\cite{iscen2019label}, and semi-supervised ML solutions like SVM~\cite{li2014towards}, and our proposed work (DL-based semi-supervised solution). 
These approaches try tackle that many successful learning algorithms need access to a large set of \emph{labeled data}. To address this issue, i.e. lack of availablity of labeled data, a combination of tri-training with a deep model wee used in~\cite{dong2018tri} to build \emph{Tri-Net}, which can use massive set of unlabeled data to help to learn with limited labeled data. The
semi-supervised deep learning model generates \textit{three} modules to exploit unlabeled data by considering model \emph{initialization}, \emph{diversity augmentation}, and \emph{pseudo-label editing}. Graph-based transduction approach that works through the propagation of few labels, called \emph{label propagation}, was used in~\cite{iscen2019label} to improve the classification performances and obtain estimated labels. This method consists of two steps. In the first step, the classifier trains through labeled and the predicts pseudo-labeled. In the second step, the nearest neighbor graph constructs based on the previous trained classifier. A limitation to this approach is that practically graph models are often mis-specified. However, this could potentially be overcomed by employing highly expressive model families like neural networks~\cite{kaiser2015neural}. Hence, in S3VM method~\cite{li2014towards}, the authors adopt SVM solution to finding the flipped label examples in a dataset and improve the safeness of the semi-supervised support vector machines (S3VM). They indicate that performance of their method is not statistically significantly worse than the solution shaped with labeled data alone. The major limitation of this method is that it is not easy to use such method for large amounts of noisy samples and outliers and it exponentially reduce ML performance.

\section{System model and proposed architecture}\label{proposedarchitecture}
In this section, we first provide a formal definition our problem (see Section~\ref{problemDefinition}). Then, in Section~\ref{architecture}, we introduce the proposed Android malware detection architecture used in the paper. In particular,  Fig.~\ref{fig:architecture} will describe the components of the proposed architecture.

\subsection{Problem definition}\label{problemDefinition}
Consider the datasets as follows.
\begin{equation}
\label{eq:eq1}\small
D=\{(x_i,y_i)\in (X,Y)\},\quad i=1,\ldots,n
\end{equation}
where $n$ is the number of malware samples. If $x_i$ has the $j$ feature, we have $x_{ij} = 1$. Otherwise $x_{ij} = 0$, and $X\subseteq \{0, 1\}^k$ - a  $k$-dimensional space. The variable $y$ represents the label of the samples with $y_i \in \{0, 1\}$ and the $D$ set has an unknown distribution on $X \times Y$. We assume the training set is defined as follows.

\begin{equation}
\label{eq:eq2}\small
S=\{(x_k,y_k)\},\quad k=1,\ldots,m
\end{equation}	
where $S$ is the label set. The flipping attack label aims to find a collection such as $P$ containing samples in $S$ so that when their labels are flipped,  it minimizes the desired target for the attacker. For simplicity, we assume that the attacker's goal is to maximize the loss function which we define it as $L(w,(x_j,y_j))$.

\subsection{Proposed architecture}\label{architecture}
\begin{figure*}[!htbp]
\centering 
\includegraphics[width=0.85\textwidth]{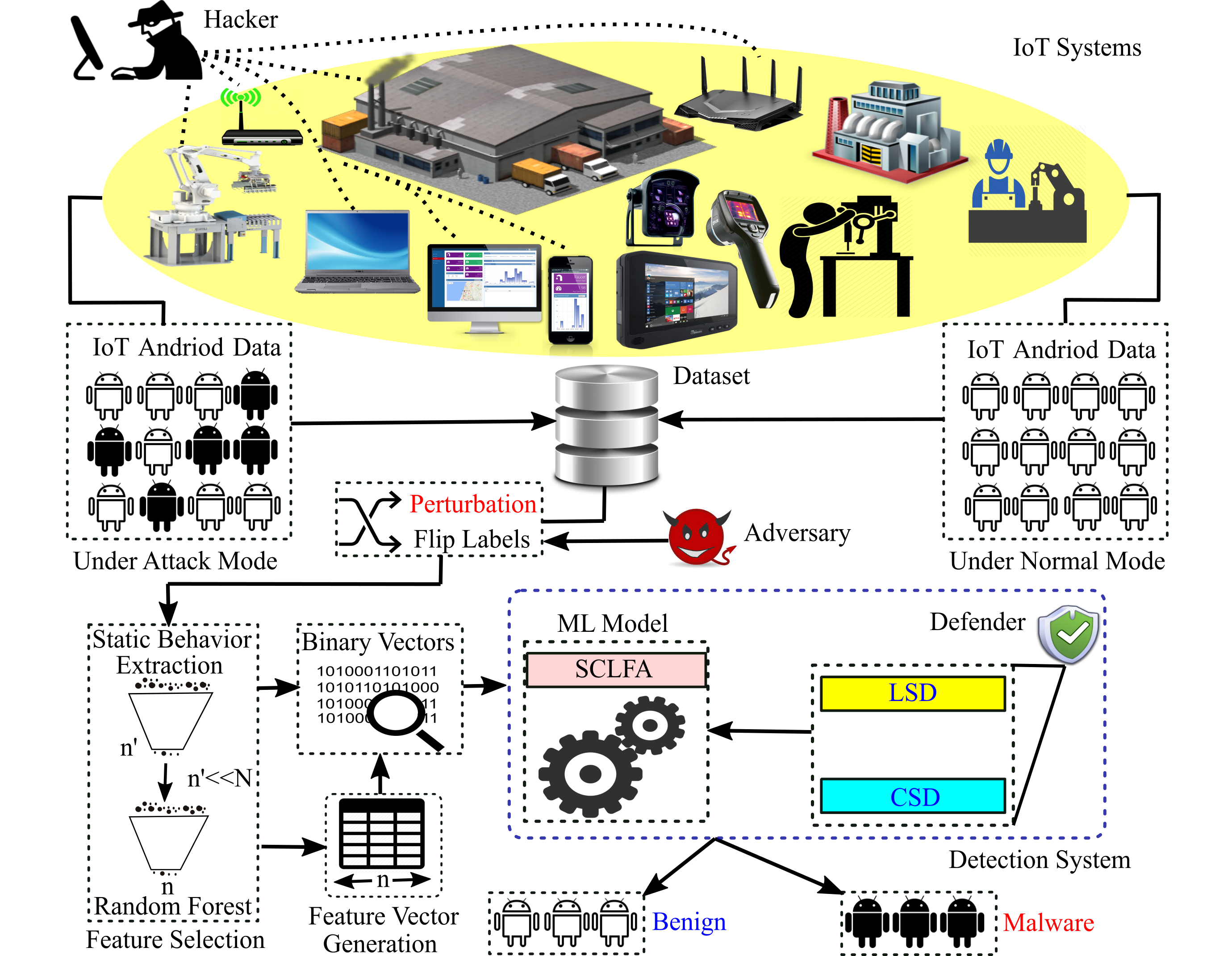}
\caption{\small Architecture overview of proposed method. ML= machine learning; SLFA is our attack method and LSD and CSD are our defense methods.}
\label{fig:architecture}
\end{figure*}

In this section, we present our architecture to tackle the Android malware detection problem in IoT systems (see Fig.~\ref{fig:architecture}).

In Fig.~\ref{fig:architecture}, we present a general scheme of our proposed architecture and the proposed attack and defense algorithms which use for Android applications. In this architecture, we assume a complex set of IoT devices (i.e., IoT systems) which are communicating with each other, represented by the yellow oval in the figure. We assume that some of the IoT devices are using Android OS platforms. We also assume that an attacker can get access to some of the IoT devices. Hence, he can manipulate the data they transferring to each other. As a result, the data traffic of each Android data can include those from malware apps, represented by the black Android app symbol in our figure.  Each Android app, whether malware and benign, presents as a vector of different features with various labels. ML algorithms exposed to adversary attacks can add a variety of perturbations to data to fool ML algorithms. Hence, in this architecture, an adversary can get access to the dataset and flip the labels by adding some perturbation of existing labels. Our feature selection component gives the ability to select the choice of features.
We then generate a binary vector of each Android app and input the result to the ML model. A final component of our architecture is the detection system composed of the ML model and our proposed defense algorithms. Our architecture can increase the robustness of our detection system against flipping label attacks, and increase the \textcolor{black}{Accuracy} of malware/benign classifications. In the following section, we explore our attack and defense algorithms.

\section{Proposed attack and defensive solutions}\label{proposedsolution}
	
In this section, the proposed classification algorithm used in the paper is described first in Section\ref{Classification}. We then describe our attack strategy, inspired by Silhouette clustering method in Section~\ref{attack}. Section~\ref{defense} presents our two defense solutions against the attack proposed in the previous section. Finally, we report the computational complexity of our strategies in Section~\ref{Complexityofalgorithms}.

\subsection{Classification algorithm}\label{Classification}

In this paper, we incorporate a \textcolor{black}{deep CNN} to classify the binary samples. We adopt the overfitting method to find out how good our dataset size is. \textit{Shift invariant} or CNN is a multilayer perceptrons strategy to tackle the fully connected neurons in each layer and help to prone the over-fitting data and can include more complex patterns. To do so, we try to classify our data using a training set and then repeat the classification using cross-validation. If we increase the data size, it gives better results in CNN classification processing.

\begin{figure*}[!htbp]
\centering 
\includegraphics[width=0.95\textwidth]{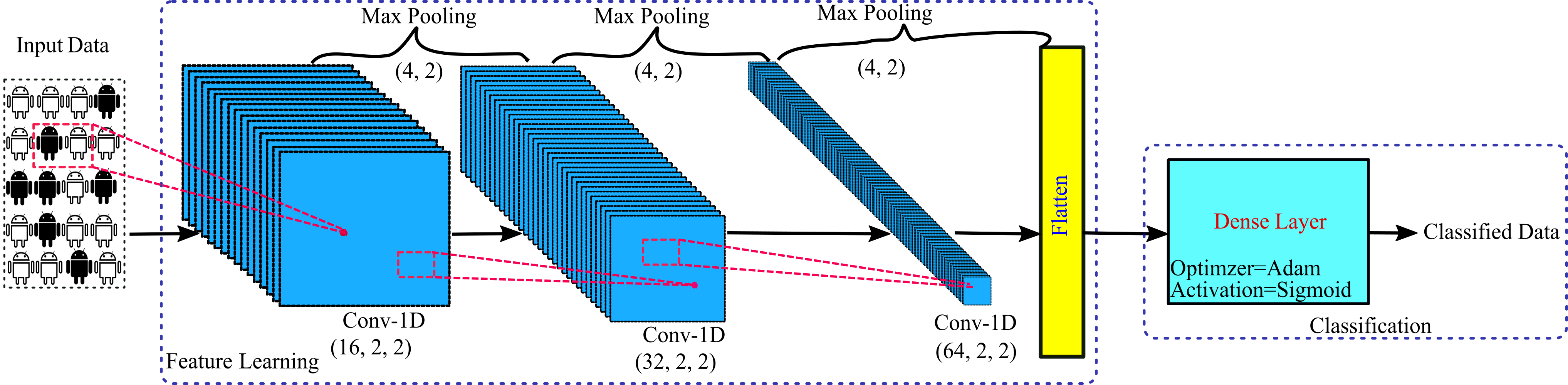}
\caption{\small Proposed classification algorithm architecture. Conv=convolution; (A,B,C)=(filters, kernel size, stride); (C,D)= (pool size, stride).}
\label{fig:fig2}
\end{figure*}

Fig.~\ref{fig:fig2} presents the proposed CNN architecture for the classification algorithm. In this figure, we can see that we apply three sequential layers of one-dimensional convolution (Conv-1D) that has 16, 32, and 64 filters. In each of these layers, we have kernel-size with value 2 and stride with value 2. We apply Maxpooling between the convolution layers to prevent overfitting by reducing the computational load, memory, and number of parameters. Each Maxpooling layer creates four pool size with two strides. After applying three convolutional layers, we adopt a Flattened layer and a Dense layer. In the Dense layer, we use  $Adam$ optimizer and $Sigmoid$ activation function to shape the classification algorithm and the out of the Dense layer is the classified data.

\subsection{Attack strategy: Silhouette Clustering-based Label Flipping Attack (SCLFA)}\label{attack}

In this subsection, we apply Silhouette clustering method to flip the labels. We name this attack  \textit{silhouette Clustering-based Label Flipping Attack (SCLFA)}. Silhouette clustering is a type of clustering technique in which can interpret and validate the consistency of data clusters. Silhouette provides a concise visual presentation object classifications. This technique defines a measurement called \textit{silhouette value (SV)} that expresses the self cluster similarity or cohesion of per object comported to other clusters or separation, which is between [-1,1]. If the silhouette value is one, it presents well matching of the object to its own cluster and is less likeness to other neighboring clusters. If the majority of the objects in a cluster have high SVs, it indicates that the clusters objects and the clustering is appropriately configured. 
We utilize a Euclidean distance method to calculate the SV in this paper. We define the \textit{label flipping attack (LFA)} as follows: 
\begin{thm}\label{labelflip}{\textbf{LFA in SCLFA:}}
LFA is a type of attack that the attacker tries to use some algorithms to modify the label of features and changes the interval of each sample in a cluster. In this paper, we use the silhouette clustering algorithm to implement LFA. To put it simply, in SCLFA, we assign an interval [-1,1] for each sample, which indicates whether the sample is in the correct cluster. If the silhouette value (SV) is negative, it means that the selected sample is a good candidate for flipping the label, and according to the silhouette algorithm, it is definitely belonging to another cluster. Hence, we change the label of such sample. Let $L_i$ be the label of the $i$-th sample out of $n$ samples in the dataset. Thus, we can write it as eq. \eqref{eq:eq3}:
\begin{equation}
\label{eq:eq3}\small
    L_i=
\begin{cases}
(x_i,y_i), & SV>0 \\
(x_i,|1-y_i|), & \text{otherwise}
\end{cases}
\end{equation}
\end{thm}
\textit{Algorithm~\ref{alg:algorithmLabelFlipping}} presents the label flipping poisoning attack.

\noindent\textbf{Description of the Algorithm~\ref{alg:algorithmLabelFlipping}.} In this algorithm, we present the proposed method, SDLFA, for the flipping label of the training sample. This method is based on the K-means clustering algorithm. In this way, we first create a model based on the K-means algorithm that divides the $X\_train$ samples into \textit{two} clusters and predicts the label for each sample (lines 1-2). Then, in line 3, we calculate the $Silhouette$ values for samples and predicted labels for the samples. As previously stated, values close to 1 indicate that the sample is fitted in the appropriate cluster, and as the values of Silhouette are less than 1 and close to -1, it means that the sample is clustered incorrectly. In the proposed method, we flipped the label of samples that have a Silhouette value less than zero. In this way, we probably have chosen the examples that have the potential to be in the other cluster (lines 4-8). 

\begin{algorithm}[H]
\caption{Silhouette-based Label Flipping Attack (SCLFA)}
\label{alg:algorithmLabelFlipping}
\textbf{Input:} X\_train , Y\_train \\
\textbf{Output:} Poisoned\_Y\_train 
 \begin{algorithmic}[1]
\color{black}
\footnotesize
\State{${M_{K}}\leftarrow$\textbf{Make Model} with two Clusters Using KMeans َAlgorithm}
\State{$Labels \leftarrow$\textbf{Predict} labels of $X\_train$ using ${M_{K}}$}
\State{$S \leftarrow$ \textbf{Compute} Silhouette values using $X\_train$ using $Labels$}

\For{each $row\in X\_train$ }
        \If{($ S[row]\leq 0$)}
            \State{$Poisoned\_Y\_train[row]=abs(1-{Y\_train}[row])$}
        \EndIf
\EndFor

\State{\textbf{return} $Poisoned\_Y\_train$}
\end{algorithmic}
\end{algorithm}

\subsection{Defensive Strategies}\label{defense}
\begin{figure}[H]
 		\centering 		\includegraphics[width=\columnwidth]{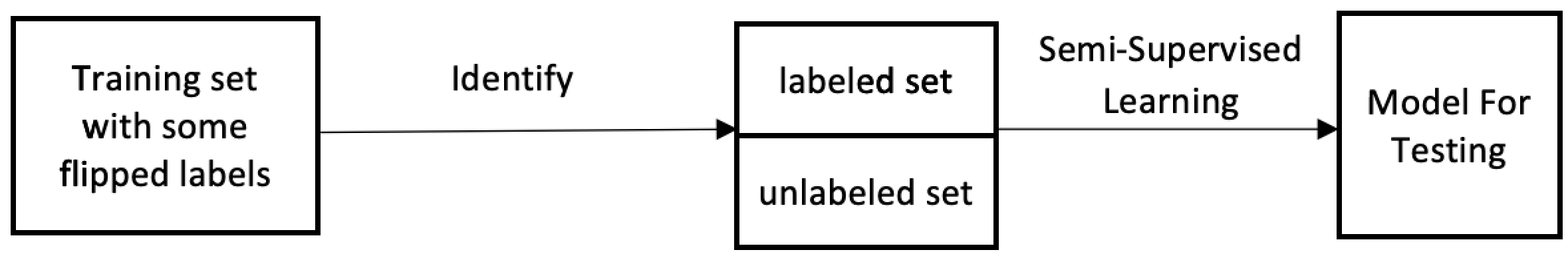}
        \caption{\small Overview of SSL model.\vspace{-10px}}
			\label{fig:fig1}
		\end{figure}

In this subsection, we discuss these countermeasures against the label flipping attack. In detail, we describe Label-based Semi-supervised Defense (LSD) and Clustering-based Semi-supervised Defense (CSD) which are presented in Sections \ref{sec:4.3.1} and \ref{sec:4.3.2}, respectively. 
In this paper, we assume our data are  only partially labeled.
Our defense strategies begins by investigating  which validation data in training samples may have been flipped. It would then predict new labels for these data and replace their labels. Fig.~\ref{fig:fig1} shows the overview of the semi-supervised learning (SSL) model for both defense strategies. 

\subsubsection{\textcolor{black}{LSD Defense }}\label{sec:4.3.1}
In this section, we design LSD algorithm to give a priority between semi-monitoring learning (SML) methods. In other words,  we adopt validation data as inputs of SML algorithms to predict the label for each sample and then rank
the predicted labels. The goal of the LSD algorithm is to find the samples for which the labels in the flipped training set are likely to have the correct values. Then, we need to give the selected data and its labels to the SSL algorithm. We need to create a validation set to monitor the training process and select the suitable parameters. That is, in the LSD method, we first rank the data points within each class, and then hold the label for the points that have the highest rankings. If no clean set is available, ranking is applied which is designed based on the multi-way classification neural network. Hence, Ranking is trained from the original training dataset. In fact, in this defense mechanism, we try to learn a binary classifier while clean labels are available. Then, we separate data containing clean labels and flipped labels. Formally speaking, in this defense strategy, in the first stage, we apply the \textit{Label Propagation (LP)} algorithm to assign the labels to unlabeled data points. Then, in the next stage, we use \textit{Label Spreading (LS)} to minimize the noises happen in labeling the samples. In LSD method, we plan to design a method which works like an ensemble learning such that it uses propagation models to predict labels for flipping. In this way, we provide a \textit{two-stage framework} for learning flipped labels. In the following, we describe LP and LS.
\begin{itemize}[leftmargin=*]
    \item \textbf{Label Propagation (LP).}
\textcolor{black}{LP is a type of semi-supervised ML algorithm can give a label to the unlabeled sample data. First, LP gives labels a small dataset of samples and make classifications. In 
other work, LP aims to propose the labels to the unlabeled
data points. That is, LP helps to find the community structure in real complex networks~\cite{aviles2019beyond}. LP compared to the other practical methods in literature has much lower processing time and could support apriori information needed about the network structure, and it does not require any knowledge of data point and samples before propagation. However, LP could produce several solutions for each set of data points.}
\item \textbf{Label Spreading (LS).}
LS algorithm is a type of propagation method that can apply the normalized graph Laplacian and soft clamping in an affinity matrix to influence on the labels. It also can diminish the regularization properties of a loss function to and make it robust against the noise~\cite{LP}. LS algorithm repeats on the modified version of a graph of data points and can normalize the edge weights by computing the normalized graph Laplacian matrix.
\end{itemize}

LP and LS algorithms create on a kernel of the system in which positively effect on the performance of the algorithm and enhance the chance of scalability of the problem. To be precise, and as an example, the RBF kernel can generate a fully connected graph that can demonstrate a dense matrix. Such big size matrix, in each iteration, could join with the cost performance full matrix multiplication calculation and results in increasing the time complexity, which causes a problem for scalable case studies. In this paper, we fix the problem by utilizing LP and LS algorithms on a KNN kernel system which provides much more memory-friendly sparse matrix and can exponentially save on execution latency. 

In the \textit{first stage} of LSD algorithm, we use validation set to train the LP and LS algorithms. Then, we use these algorithms to predict labels of training set. At the same time we train the CNN classifier with the validation data and predict new labels for training set samples. In the second stage, we use voting between all available labels, i.e., LP output, Label Spreading, CNN predicted labels and poisoned labels.  

In the \textit{second stage} of LSD algorithm, we apply a temporary ensembling for semi-supervised Deep Neural Network training. Then, we present a semi-supervised two-stage algorithm for training flipped labels, which include two main components. We discover and select some samples from the labeled training set, for which there are strong indications that their labels are correct. Afterward, we aim to learn a semi-supervised deep neural network that only uses the selected labels from the first previous stage. Finally, the ML model network can easily classify previously unseen test data. We summarize the proposed LSD countermeasure algorithm in \emph{Algorithm~\ref{alg:LSDsemisupervised}}.

\noindent\textbf{Description of the Algorithm~\ref{alg:LSDsemisupervised}.} It presents the semi-supervised defense, which is based on Label estimation. As seen in this algorithm, in lines 3-5, the label spreading algorithm is applied, which is used to find labels of training data. The label spreading algorithm is trained using validation data and then created a model is used to predict labels of training data. Similarly, lines 6-8 use the Label Propagation algorithm to predict training data labels. This algorithm, like the Label Spreading algorithm, is a semi-supervised algorithm. In lines 9 and 10 of this algorithm, convolutional neural network as the third part of the ensemble learning approach is used, which is trained with validation data and is used to predict the training data label. The final part of the LSD method is the voting between the results of the three methods described and the poisoned label, which is the result of voting as the label for training samples. 

\begin{algorithm}[H]
\caption{Label-based Semi-supervised Defense (LSD)}
\label{alg:LSDsemisupervised}
\textbf{Input:} $X\_validation , Y\_validation , X\_train , Poisoned\_Y\_train$\\
\textbf{Output: ${Y_{Corrected}}$} 
 \begin{algorithmic}[1]
\color{black}
\footnotesize
\State{$X\leftarrow  X\_train$}
\State{$Y\leftarrow  Y\_validation$}
\State{${M_s}\leftarrow$\textbf{Make Model} using LS algorithm}
\State{\textbf{Fit} the ${M_s}$ Model on $X$ and $Y$}
\State{${L_s} \leftarrow$ \textbf{Predict} labels of $X\_train$ using ${M_s}$}
\State{${M_p}\leftarrow$ \textbf{Make Model} using LP algorithm}
\State{\textbf{Fit} the ${M_p}$ Model on $X$ and $Y$}
\State{${L_p} \leftarrow$ \textbf{Predict} labels of $X\_train$ using ${M_p}$}
\State{${M_{cnn}}\leftarrow$ \textbf{Make Model} using proposed CNN algorithm}
\State{${L_{cnn}} \leftarrow$ \textbf{Predict} labels of $X\_train$ using ${M_{cnn}}$}
\State{${Y_{Corrected}}=\textbf{Voting}({Y_{Corrected}},{L_s},{L_p},{L_{cnn}})$}

\State{\textbf{return ${Y_{Corrected}}$} }
\end{algorithmic}
\end{algorithm}

\subsubsection{\textcolor{black}{CSD Defense}}\label{sec:4.3.2}
The main idea behind this approach is to use clustering techniques to correct flipped labels. As each of the clustering methods has its specific measure, in this method it is suggested to use the voting between the label determined by different clustering methods for determines the label of the flipped samples. Hence, we use four indices to analyze the \textcolor{black}{Accuracy} of our generated clusters and the predicted one and identify the most likely adversarial examples and flip their labels. 

\noindent\textbf{Description of Algorithm~\ref{alg:algorithmCSD}.} In this algorithm, we explain the CSD method. In lines 1-3 of this algorithm, we use the proposed CNN model and validation data and predict the labels of the training data. Lines 4 to 7, the algorithm describes four cluster metrics, namely RI, MI, HM, and FMI and compute their values. Each of these metrics is a measure for the \textcolor{black}{Accuracy} of clustering. The main idea behind this approach is that the training samples are labeled in such a way that the mentioned measure does not differ significantly from the values calculated from the validation data. Therefore, in lines 8-16, we add one sample of the training data to the validation dataset, calculate the values of the clustering with four metrics , and compare them with the base values. If the difference is less than 0.1 (i.e., we consider as a threshold), then we consider the sample to be properly labeled. As a result, the output of this algorithm is the labeled sample which can be used as a validation data and selected sample for training the ML model.
\begin{algorithm}[H]
\caption{Clustering-based Semi-supervised Defense (CSD)}
\label{alg:algorithmCSD}
\textbf{Input:} $X\_validation , Y\_validation , X\_train , Poisoned\_Y\_train$\\
\textbf{Output: ${Y_{Corrected}}$} 
 
\begin{algorithmic}[1]
\color{black}
\footnotesize
\State{$X\leftarrow X\_validation$}
\State{${M_{cnn}}\leftarrow$ Make Model Using proposed convolutional neural network}
\State{${Y_{Corrected}} \leftarrow$ predict labels of $X\_train$ using ${M_{cnn}}$}
\State{$R\leftarrow$Compute \textbf{Rand Index} using $X$ and ${Y_{Corrected}}$}
\State{$M\leftarrow$Compute \textbf{Mutual Information} using $X$ and ${Y_{Corrected}}$}
\State{$H\leftarrow$Compute \textbf{Homogeneity Metric} using $X$ and ${Y_{Corrected}}$}
\State{$F\leftarrow$Compute \textbf{Fowlkes-Mallows Index} using $X$ and ${Y_{Corrected}}$}

\For{each $row$ $\in$ $ X\_train$}
\State{$X\_temp\leftarrow X\_validation+row$}
\State{Compute $R_{temp}$ ,$M_{temp}$ ,$H_{temp}$ ,$F_{temp}$}
\State{$S\leftarrow |((R_{temp}-R)+(M_{temp}-M)+(H_{temp}-H)+(F_{temp}-F))|$}
\If{($S\leq 0.1$)}
	\State{$X\leftarrow X+row$}
	\State{${Y_{Corrected}} \leftarrow {Y_{Corrected}}$+Label related to $row$}
\EndIf
\EndFor
\State{\textbf{return ${Y_{Corrected}}$} }

\end{algorithmic}
\end{algorithm}
The indices are defined as below.
\begin{itemize}[leftmargin=*]
 \item  \textbf{Rand Index (RI):}
 Rand measure/index is a statistical index to calculate the similarity between two data clustering~\cite{rand1971objective}. It is a value between zero and one such that zero indicating that two sets of clustered data do not have any pair point and one indicating that the data clustering is the same. Also, RI can be used to adjust a group for elements that we called them \textit{adjusted Rand index}. In other words, RI is a metric of the \textcolor{black}{Accuracy} of two sets of data points, which represents the frequency of occurrence of total pairs. Formally, speaking, RI presents the probability of how can we randomly select two pair $X_1$ and $X_2$ in two partitions of the same big set.

 \item  \textbf{Mutual Information (MI):}
  MI, or \textit{information gain} is a measure to realize the amount of information and dependency between two separate variables by observing them~\cite{MI}. It is a type of entropy of a random variable that can understand the joint distribution of a pair data points which calculates by the product of the marginal distribution of those pair samples. Since the data we are dealing with are fallen in the group of discrete data with discrete distribution, we can calculate the $\mathcal{I}$ MI of two jointly 
  discrete random variable $X_1$ and $X_2$ as follows:
\begin{equation}\label{eq:eq001}\small
\begin{split}
\mathcal{I}(X_1,X2)=\sum_{x_1\in X_1}\sum_{x_2\in X_2} p_{(X_1,X_2)}(x_1,x_2)\log_2\\ \left(\frac{p_{(X_1,X_2)}(x_1,x_2)}{p_{X_1}(x_1)p_{X_2}(x_2)}\right)
\end{split}
\end{equation}
where $p_{(X_1,X_2)}$ is a joint probability mass function for the two samples of $X_1$ and $X_2$ and $p_{X_1}$ is a marginal probability of sample $X_1$ and $p_{X_2}$ is a marginal probability of sample $X_2$.

 \item  \textbf{Homogeneity Metric (HM):}
  This metric uses for validating the data points which are members of a single class. HM is independent of being changed the score value of data point when a permutation of the class or labels are applied~\cite{hirakawa2005adaptive}. We can define HM values as $\mathcal{HM}$ as follows:
  \begin{equation}\label{eq:eq002}\small
\mathcal{HM}=1-\frac{H(Y_{T}|Y_{PR})}{H(Y_{T})}
\end{equation}
 where $\mathcal{HM}$ can be between 0 and 1. Note that low values of $\mathcal{HM}$ explains a low homogeneity, vice-versa. If we have a sample data $Y$, we define $Y_{PR}$, $Y_{T}$ are the predicted and the corrected value for that sample, hence, $H(Y_{T})$ is HM value for that sample when it is correctly placed and predicted to be placed in one single class, respectively. Besides, the $\frac{H(Y_{T}|Y_{PR})}{H(Y_{T})}$ indicates that the predicted sample is not placed correctly in a single class. We aim to approach this fraction smaller and reach it to zero $(\mathcal{HM}\rightarrow 1)$. We can achieve this goal when we reduce the knowledge of $Y_{PR}$ and diminish the uncertainty of $Y_{T}$ that results in the fraction above become smaller, and we have HM around 1.

 \item  \textbf{Fowlkes-Mallows Index (FMI):}
 Fowlkes–Mallows Index (FMI) metric is a popular metric to understand the similarity between two generated clusters, whether hierarchical or benchmark classification clusters~\cite{guo2019clustering}. The higher similarity between two clusters (created cluster and the benchmark one) indicates higher FMI values. FMI is an accurate metric uses to evaluate the unrelated data and also is reliable even with added noises to the data results. 
\end{itemize}

\subsection{Computational Complexity}\label{Complexityofalgorithms}

In following section, we evaluate computational complexity analysis on the presented attack and defensive methods. Assume that the number of samples in $X\_train$ and $X\_validation$ are $n$ and $m$, respectively. We list the computational complexity of the methods. So, we have

\begin{itemize}[leftmargin=*]
  \item \textbf{\textcolor{black}{Time Complexity of SCLFA Attack}}\\
  \textcolor{black}{Focusing on SCLFA, the computation of all possible configurations in line 1-2 of Algorithm~\ref{alg:algorithmLabelFlipping} creates a model based on the $K$-means method and predicts the correct $n$ training samples, results in  $\mathcal {O} (n ^ 2\times k) $. Since in this method, $k = 2$, the time complexity is in the order of $ \mathcal {O} (n ^ 2) $. In line 3 of this algorithm, Silhouette values are computed for $n$ training data samples, which has a complexity of  $ \mathcal {O} (n ^ 2) $. Lines 4-8 of the algorithm include a \textit{for} loop that performs the correction of the $m$ validating labels and has a complexity of $ \mathcal {O} (m) $. Overall, the computational complexity of the Algorithm~\ref{alg:algorithmLabelFlipping} is in the order of
  $\mathcal {O} (n ^ 2) +  \mathcal {O} (n ^ 2) +\mathcal{O}(m)$=$\mathcal{O}(n ^ 2)$, $\forall\ n \gg m$.}\\
  
  \item \textbf{\textcolor{black}{Time Complexity of LSD Defense}}\\ 
  \textcolor{black}{Focusing on LSD, the computation of  Algorithm~\ref{alg:LSDsemisupervised} directly relates to the LS method, which has a complexity of $\mathcal {O} (n) $. Similarly, in lines 6-8, the model is based on the LP algorithm, which has a complexity of $ \mathcal {O} (n) $. Then, lines 9 and 10 present CNN model creating, according to~\cite{he2015convolutional}, which has a computational complexity of all convolutional layers. CNN computational complexity is 
  $ \mathcal {O} (\Sigma {^d}_{i=1} {n_{(l-1)}}\times s^2_l \times m^2_l) $, where $l$ is the index of a convolutional layer; $d$ is the depth (number of convolutional layers); $n_l$ is the width or the number of filters in the $l_{th}$ layer--$n_{(l−1)}$ is the number of input channels of the $l$-th layer; $s_l$ is the spatial size (length) of the filter and $m_l$ is the spatial size of the output feature of CNN which has a  time complexity in the order of $ \mathcal {O} (n ^ 3)$.
Then, we performs voting between results that has a complexity of $ \mathcal {O} (1)$ (line 11). Overall, the computational complexity of LSD defense algorithm is
  $\mathcal{O}(n)+\mathcal{O}(n)+\mathcal {O} (n ^ 3)+\mathcal {O} (1)$=$\mathcal{O}(n ^ 3)$.}\\

  \item \textbf{\textcolor{black}{Time Complexity of CSD Defense}}\\
 \textcolor{black}{Focusing on CSD, the computation of  Algorithm~\ref{alg:algorithmCSD} relies on CNN model construction based validation data (lines 1-2). Then, we predict the label for training data samples based on this generated ML model. Therefore, the computational complexity of this part is in the order of $ \mathcal {O} (n ^ 3) $. Focusing on the RI, MI , HM and FMI clustering metric calculations, they have a complexity of $\mathcal{O}(n)$ (lines 4-7). Then, we calculate the values of these parameters for $m$ samples. Hence, the complexity of this loop of the CSD algorithm is in the order of $ \mathcal {O} (n\times m) $ (lines 8-16). As a result, the overall computational complexity of CSD defense method is 
  $\mathcal{O}(n \times m)+\mathcal{O}(n ^ 3) +\mathcal{O}(n )$=$\mathcal{O}(n^3)$,$\forall\ n>>m$.}\\
  \end{itemize}


\section{Experimental evaluation}\label{resultAnalysis}
In this section, we report the results of  our proposed attack and defense algorithms in different scenarios: with feature selection consideration (WFS) and without feature selection consideration (WoFS). Given the two scenarios, we conduct our experiments on our attack (SCLFA) and defense algorithms (LSD and CSD) against KNN-based Semi-Supervised Defense (KSSD)~\cite{paudice2018label}. The source code of the paper is available in Github~\cite{Teheri2020NCAA}.


\subsection{Simulation setup}\label{SimulationSetup}
We describe the test metrics, datasets, features, classification parameter, and comparison defense algorithm below.

\subsubsection{Test metrics}\label{sec:5.1.1}
To provide a comprehensive evaluations of our attack and defense algorithms, we use the following indices: \textcolor{black}{Accuracy, Precision, Recall, False positive rate (FPR), True negative rate (TNR), miss rate (FNR), F1-score}, and area under cover (AUC):
\begin{itemize}[leftmargin=*]

\item\textbf{Accuracy:} Accuracy metric is defined in:
\begin{equation}
\label{eq:eq16}\small
Acc=\frac{\Omega+\chi}{\Omega+\chi+\Lambda+\nu}
\end{equation}
where $\Omega$ is true positive; $\chi$ is true negative; $\Lambda$ is false positive, and $\nu$ is false negative metrics.  

\item\textbf{Precision:} Precision is the fraction of relevant samples between the retrieved samples which is shown in
\begin{equation}
\label{eq:eq17}\small
Precision=\frac{\Omega}{\Omega+\Lambda}
\end{equation}

\item\textbf{Recall:} The Recall is expressed in
\begin{equation}
\label{eq:eq18}\small
Recall=\frac{\Omega}{\Omega+\nu}
\end{equation}
\item \textbf{\textcolor{black}{F1-Score:}} This metric defines as a harmonic mean of \textcolor{black}{Precision and Recall} which is defined as
\begin{equation}
\label{eq:eq182}\small
\begin{split}
\textcolor{black}{F1-Score}=\frac{1}{\frac{1}{Recall}+\frac{1}{Precision}}=\\2*\frac{Precision*Recall}{Precision+Recall}
\end{split}
\end{equation}

\item\textbf{False Positive Rate (FPR):} This metric represents a ratio between the number of negative events incorrectly classified as positive (false positives) and the total number of actual negative events. This metric is described in equation \eqref{eq:eq19}:
\begin{equation}
\label{eq:eq19}\small
FPR=\frac{\Lambda}{\Lambda+\chi}
\end{equation}
\item\textbf{Area Under Curve (AUC):} AUC measures the trade off between misclassification rate and FPR. This metrics can be calculated as \eqref{eq:eq20}:
 \begin{equation}
\label{eq:eq20}\small
AUC=\frac{1}{2}\left(\frac{\Omega}{\Omega+\Lambda}+\frac{\chi}{\chi+\Lambda}\right)
\end{equation}
\item\textbf{False Negative Rate (FNR):} This metric is a method for determining the case that the condition does not hold, while in fact it does. In this work, we also called it \textit{miss rate}. This metrics can be calculated as \eqref{eq:eq21}:
 \begin{equation}
\label{eq:eq21}\small
FNR=\frac{\nu}{\nu+\Omega}
\end{equation}
\end{itemize}

\subsubsection{Datasets}\label{sec:5.1.2}
Our experiments utilized the following \textit{three} datasets:
\begin{itemize}[leftmargin=*]
\item \textit{Drebin dataset:} This dataset is an Android example collection that we can apply directly. The Drebin dataset includes 118,505 applications/samples from various Android sources~\cite{arp2014drebin}. 

\item \textit{Contagio dataset:} it consists of 11,960 mobile malware samples and 16,800 benign samples~\cite{Contagio}.

\item \textit{Genome dataset:} This dataset is an Android example which is supported by the National Science Foundation (NSF) project of the United States. From August 2010 to October 2011, the authors collected about 1,200 samples of Android malware from different categories as a \textit{genome} dataset~\cite{jiang2012dissecting}.  

\end{itemize}

\subsubsection{Features}\label{sec:5.1.3}
In this paper, we consider various malicious sample features like Permissions, APIs and Intents. We summarize them as follows:
\begin{itemize}[leftmargin=*]
    \item \textit{Permission:} Permission is a essential profile of an Android  application (apk) file that includes information about the application. The Android operating system processes these Permission files before installation.  
    
    \item \textit{API:} API feature monitors various calls to APIs on an Android OS, such as sending SMS or accessing a user's location.
    
    \item \textit{Intent:} Intent feature applies to represent the communication between different components which is known as a \textit{medium}.
\end{itemize}
\subsubsection{Parameter setting}\label{sec:5.1.4}
We rank the features to better manage the huge amount of features using the \textit{RandomForestRegressor} algorithm. Then, we repeat our experiments for 300 manifest features with higher ranks to determine the optimal number of features for modification in each method. In each test, we randomly consider 60\% of the dataset as training samples, 20\% as validation samples and 20\% as testing samples. We run our experiments on an 8-core Intel Core i7 with speed 4 GHz with 16 GB RAM on an OS Win10 64-bit.

\subsubsection{Comparison of defense algorithms}\label{sec:5.1.5}
 \textcolor{black}{We compare our proposed algorithms to defend against label flipping attacks with KNN-based Semi-Supervised Defense (KSSD)~\cite{paudice2018label} and GAN-based Defense~\cite{taheri2019can}.} The comparison results show that our proposed methods are more robust in detecting Flipping Label attacks. In the KSSD method, authors adopt $K$ nearest neighbor (KNN) method to mitigate the effect of label flipping attacks. A relabeling mechanism for suspected malicious malware is suggested. The KNN algorithm uses the training set to assign a label to each sample. The aim is to ensure the homogeneity of the label between the close examples, especially in areas that are far from the decision boundary. In the training set, authors first select $K$ nearest neighbors using the Euclidean distance. Then, if the fraction of the data points that are among the most commonly enclosed labels in $K$ are equal or greater than the threshold of $t$ with $0.5\leq t\leq 1$ they select them. The training sample available in the $K$ nearest neighbor is relabeled with the most common label. Given that we only have two types of labels in detecting malware, they assign the dominant label in $K$ to the nearest neighbor to the sample. \textit{Algorithm ~\ref{alg:algorithmKNNdefense}} presents the KSSD defense.

\begin{algorithm}[H]
\caption{KNN-based Semi-Supervised Defense (KSSD)~\cite{paudice2018label}}
\label{alg:algorithmKNNdefense}
\textbf{Input:} training set $S$, Threshold $t$ \\
\textbf{Output:} $S$
 \begin{algorithmic}[1]
\color{black}
\footnotesize
\For{$i=1\leq m$}
\State{$K_{NN}\longleftarrow$\textbf{Find} $K$ nearest neighbor for sample $i$}
\State{$F\longleftarrow$ \textbf{Find} the fraction of samples in KNN with most frequency}
\If{$F>t$}
\State{$y_i$= label with most frequency in $K_{NN}$}
\EndIf
\EndFor
\State{\textbf{return} $S$}
\end{algorithmic}
\end{algorithm}

We indicate that poisoning sample points that are far from the decision boundary are likely to be relabeled, and reduce the negative performance consequences on the classification algorithm. Although the algorithm gains validation of genuine points at the same time, i.e., in areas where the two classes overlap (especially for values of t close to 0.5), we can have a similar amount of the correct points that are labeled in two classes, and it confirms that the KSSD label correction solutions presented in Algorithm~\ref{alg:algorithmKNNdefense} must be the same for the two classes. Therefore, this type labeling shall not considerably influenced the classification algorithm.\\
\textcolor{black}{Another comparison made in this paper is the GAN-based defense presented in ~\cite{taheri2019can}. Algorithm ~\ref{alg:algorithmDefGAN} illustrates the proposed method in this study. This algorithm works by generating new samples to train the machine learning model again. Specifically, in this paper, we use the GAN as a synthetic data generator set. GAN has two functions called \textit{Generator} and \textit{Discriminator}. The former one  can modify the less likely malware samples. To do so, in the training phase, it selects one random feature from the highest ranked features with zero value. Then, it changes the selected feature value to one to generate new sample. In the latter function, the GAN use this function as a classifier to predict the class variable. It  modifies the features until the discriminator function is cheated and labels such a sample  among the benign samples. Besides, we gather the wrongly estimated malware samples into a synthetic data generator set. Besides, we use 80\% of the synthetic data generator set with the training dataset to update the AML model. We use the remaining synthetic data generator samples (i.e., 20\% of the data samples) with the test dataset to test/analyze the  classification. It is found that the proposed methods even outperform the GAN-based method, since the proposed GAN is only flipping-focused research with respect to the important features of decision making, while the proposed methods in this paper are based on the value of labels.}

\begin{algorithm}[H]
\caption{\small GAN: pseudo-code of the GAN defense ~\cite{taheri2019can}}
\label{alg:algorithmDefGAN}
\textbf{Input:} training set $S$($X$, $Y$), $X^*$, $Y^*$, $\lambda$ \\
\textbf{Output:} $Model_{new}$
\begin{algorithmic}[1]
\footnotesize
\State{$Model_{poison}\leftarrow$ Fit a model on $X_{trn}$ using $LR$}
\State{$Lesslikely\leftarrow$ Fit 10\% model of $X_{m}$ with KNN to $Model_{poison}$}
\For{each $x\in Lesslikely$ }
\State{$x_{new}\leftarrow x$}

\While{$x_{new} \in Model_{poison}$ Classify as $M$}
\State{$x_{new}\leftarrow$ Add ranked{($\lambda$)} features from $X_{b}\cup\ x_{new}$}
\EndWhile
\State{$synthetic_{data}\leftarrow synthetic_{data}\cup x_{new}$}

\EndFor
\State{$Model_{new}\leftarrow$ Fit Model on $X_{trn}\cup synthetic_{data}$}
\State{$Y_{corrected}\leftarrow$ Use $Model_{new}$ to predict label of X }
\State{$ُS\leftarrow$X and related label($Y_{corrected})$}
 \State{\textbf{return} $S$}
\end{algorithmic}
\end{algorithm}

 \subsection{Experimental results}\label{results}
 
In this section, we test our presented attack algorithm (SCLFA) on our originally trained classifiers and validate our defenses algorithms (LSD and CSD) against adversarial label flipped examples (KSSD) \textcolor{black}{and GAN-based synthetic data generator} on three above Android malware datasets.

\subsubsection{Comparing methods based on Precision, Recall, F1-score}\label{sec:5.b.1}
\begin{figure*}[!htb]
    \centering
	\begin{subfigure}{0.32\textwidth}
 		\centering 
 		\includegraphics[width=1.01\linewidth]{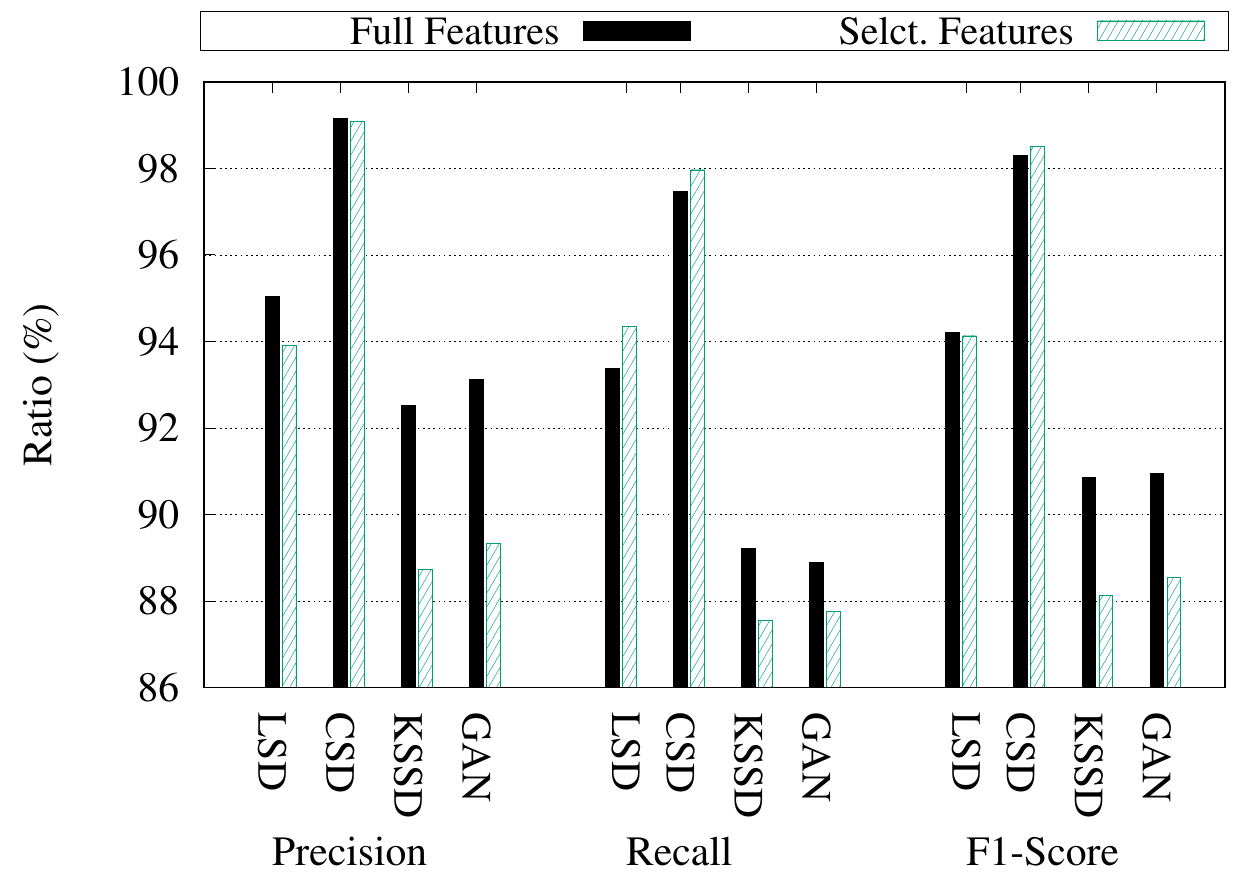}
			\caption{\small Drebin-API}
			\label{fig:fig10a}
 	\end{subfigure} 
   \begin{subfigure}{0.32\textwidth}
 		\centering
 		\includegraphics[width=1.01\linewidth]{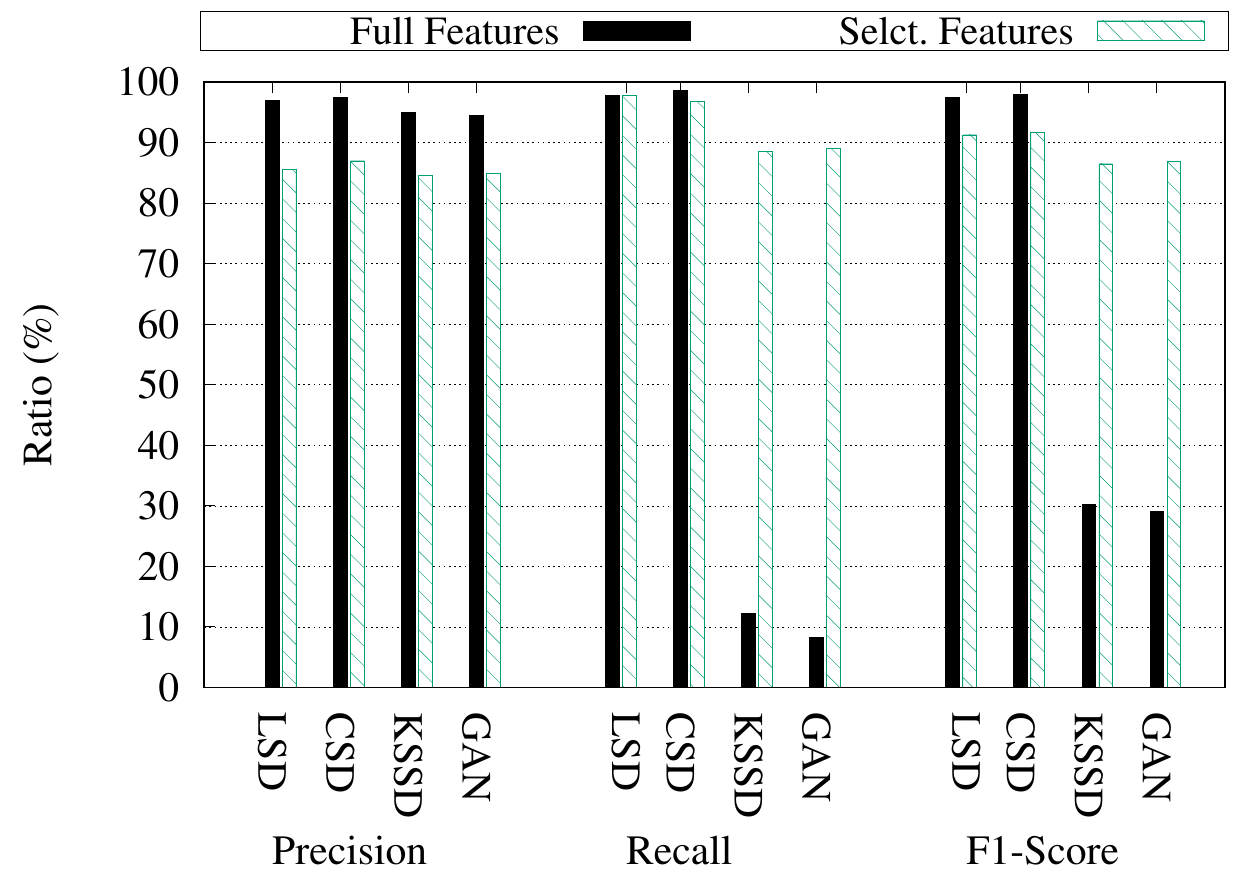}
			\caption{\small Drebin-permission}
			\label{fig:fig10b}
 	\end{subfigure} 
    \begin{subfigure}{0.32\textwidth}
 		\centering 
 		\includegraphics[width=1.01\linewidth]{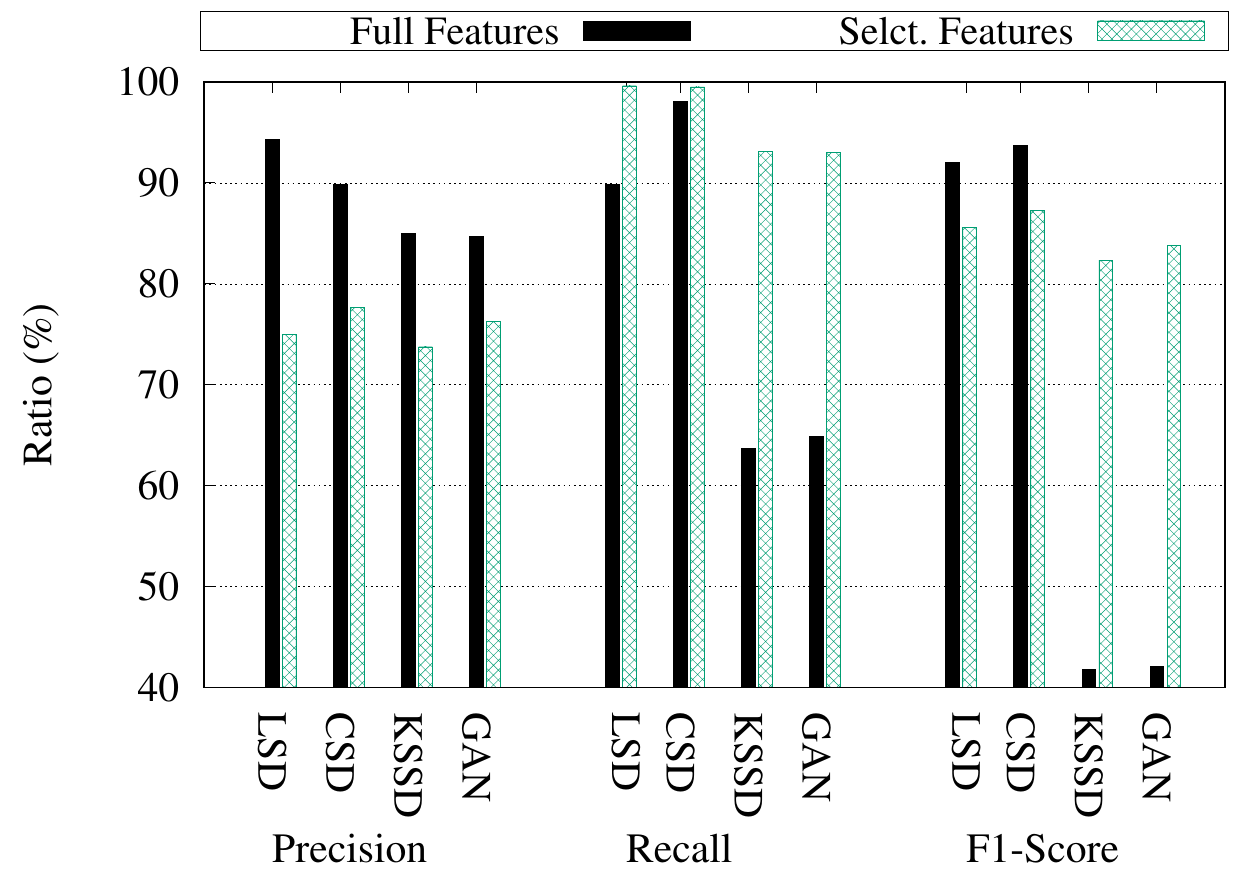}
		\caption{\small Drebin-intent}
			\label{fig:fig10c}
 	\end{subfigure}
 		\hfill
 	\begin{subfigure}{0.32\textwidth}
 		\centering
 		\includegraphics[width=1.01\linewidth]{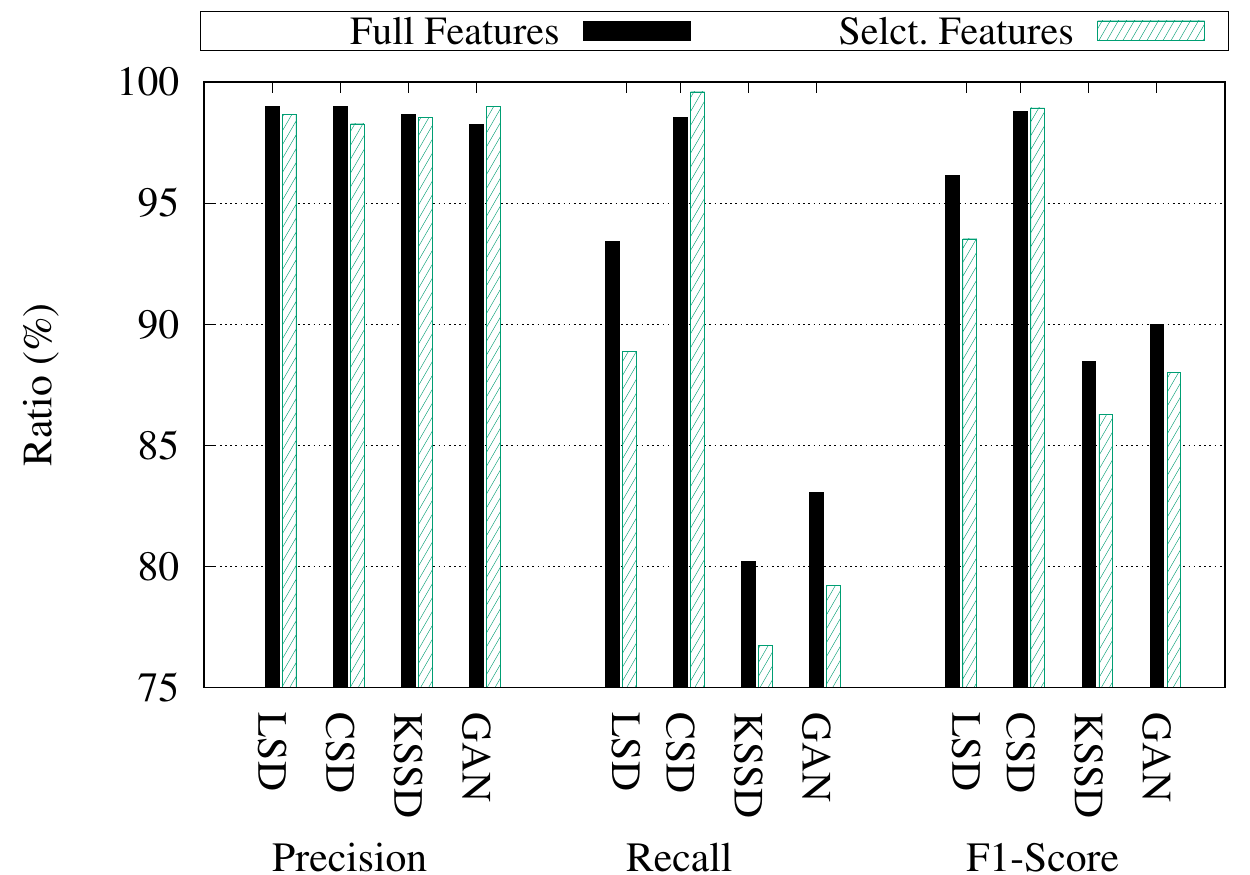}
			\caption{\small Contagio-API}
			\label{fig:fig10d}
 	\end{subfigure}  
    \begin{subfigure}{0.32\textwidth}
 		\centering 
 		\includegraphics[width=1.01\linewidth]{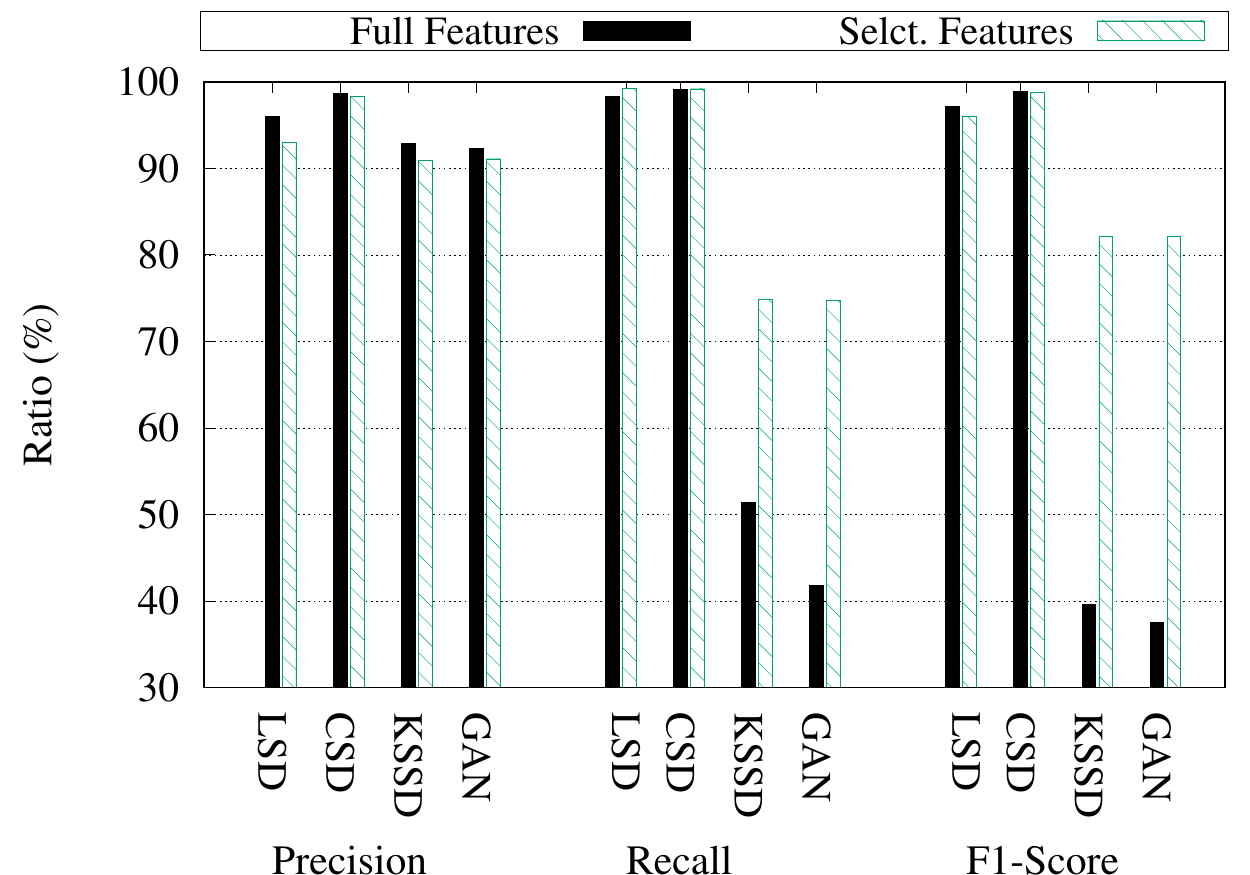}
			\caption{\small Contagio-permission}
			\label{fig:fig10e}
 	\end{subfigure}
   \begin{subfigure}{0.32\textwidth}
 		\centering
 		\includegraphics[width=1.01\linewidth]{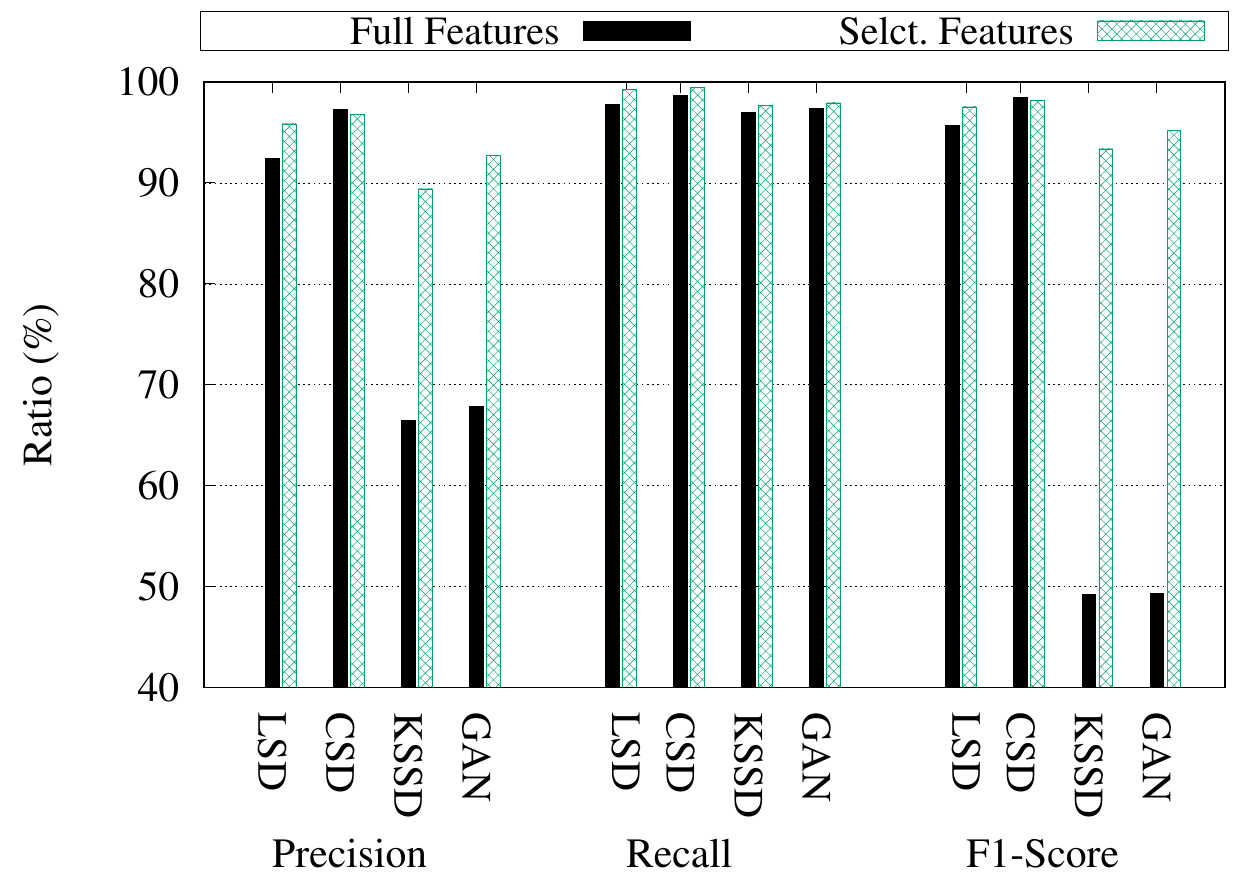}
		\caption{\small Contagio-intent}
			\label{fig:fig10f}
 	\end{subfigure} 
 	\begin{subfigure}{0.32\textwidth}
 		\centering
 		\includegraphics[width=1.01\linewidth]{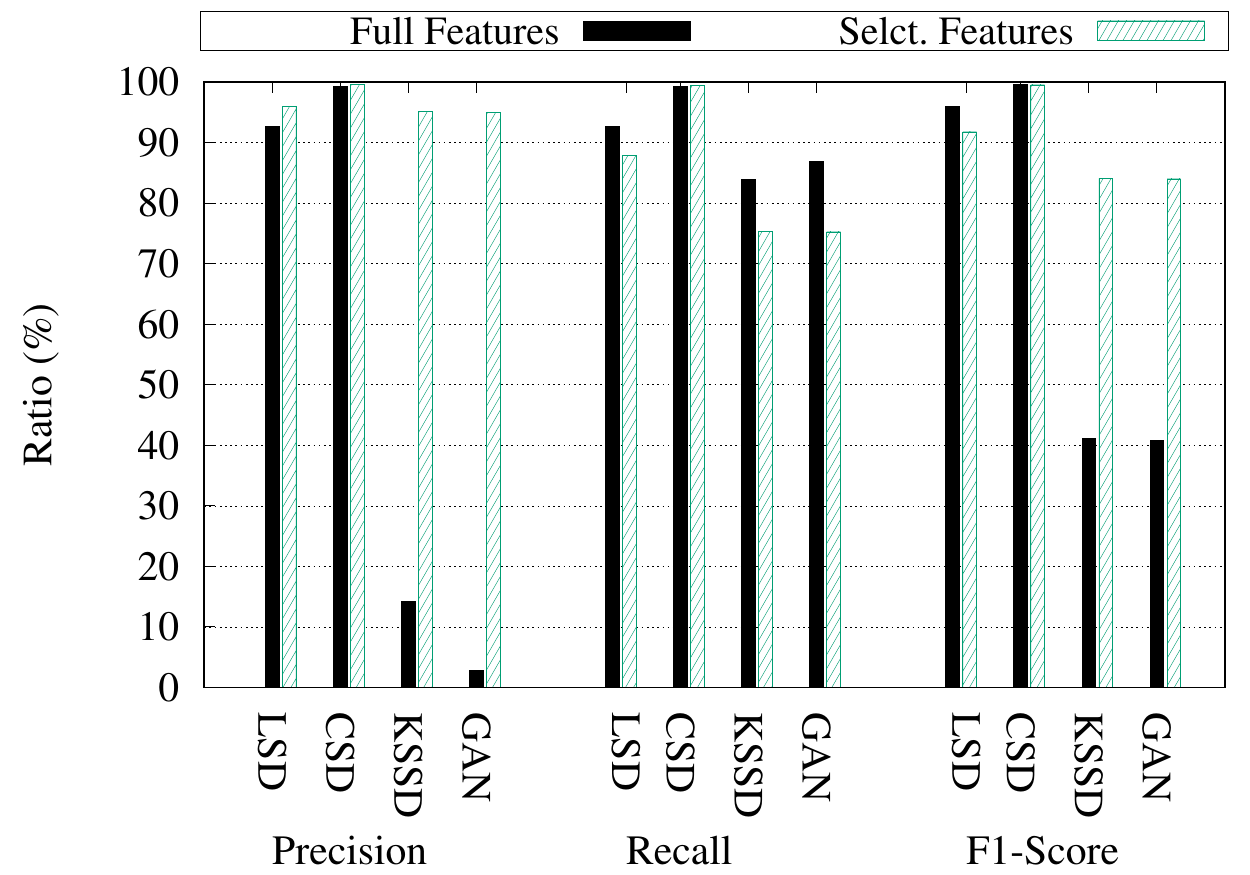}
			\caption{\small Genome-API}
			\label{fig:fig10g}
 	\end{subfigure}  
    \begin{subfigure}{0.32\textwidth}
 		\centering 
 		\includegraphics[width=1.01\linewidth]{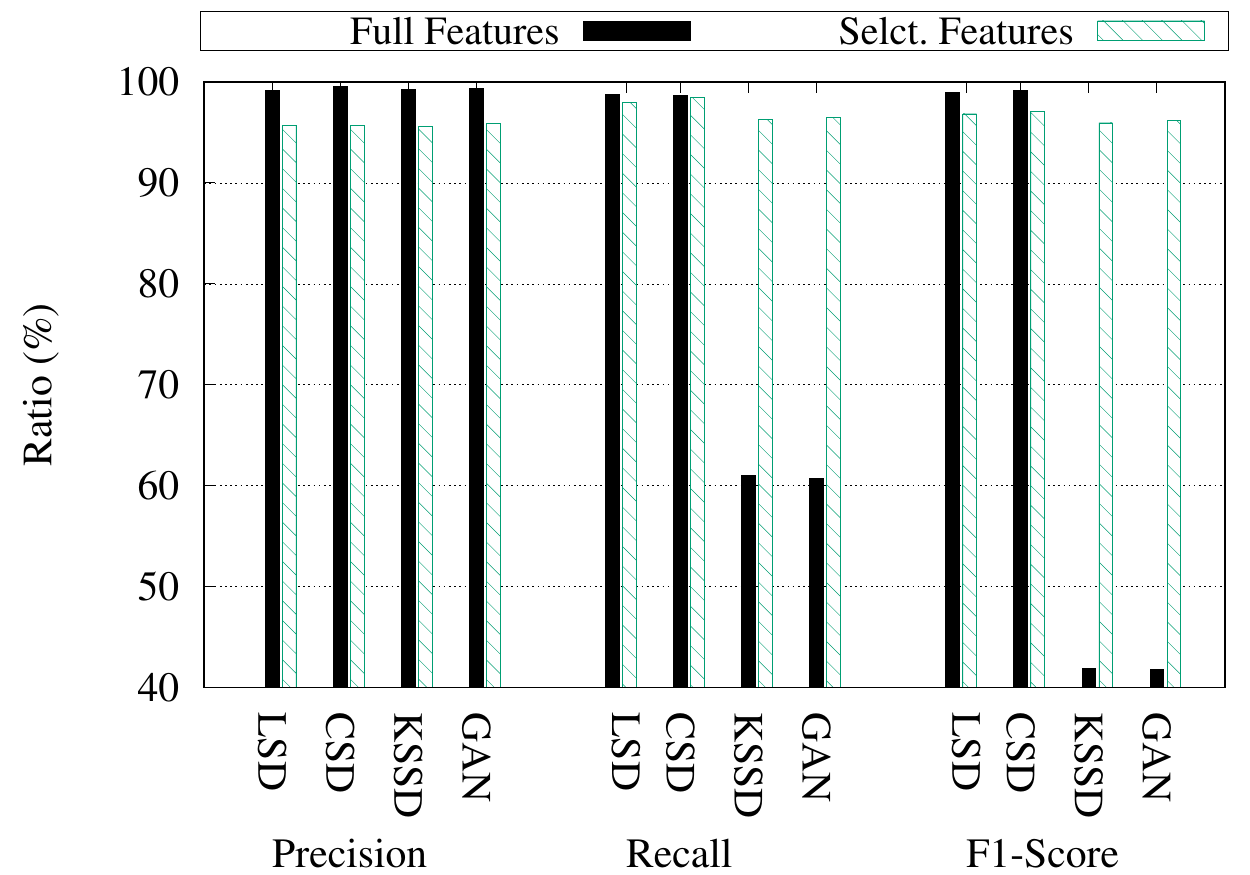}
			\caption{\small Genome-permission}
			\label{fig:fig10h}
 	\end{subfigure}
   \begin{subfigure}{0.32\textwidth}
 		\centering
 		\includegraphics[width=1.01\linewidth]{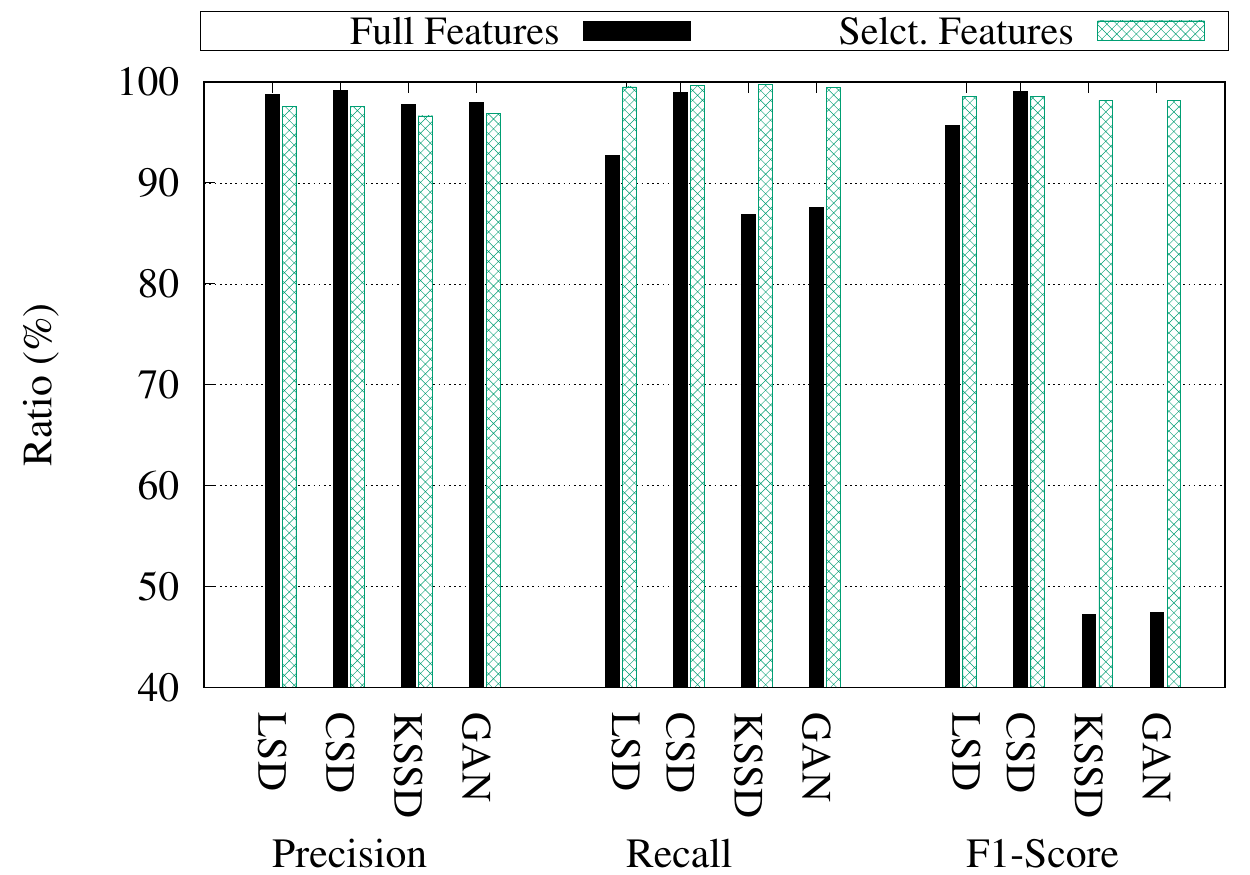}
		\caption{\small Genome-intent}
			\label{fig:fig10i}
 	\end{subfigure} 
 	\hfill
        \caption{\small Comparison between DEFENSE algorithms with reference to \textcolor{black}{Precision, Recall, and F1-Score} for API, Intent and Permission features in various datasets.\vspace{-10px}}
			\label{fig:fig10}
		\end{figure*}

\begin{figure*}[!htb]
    \centering
	\begin{subfigure}{0.32\textwidth}
 		\centering 
 		\includegraphics[width=1.01\linewidth]{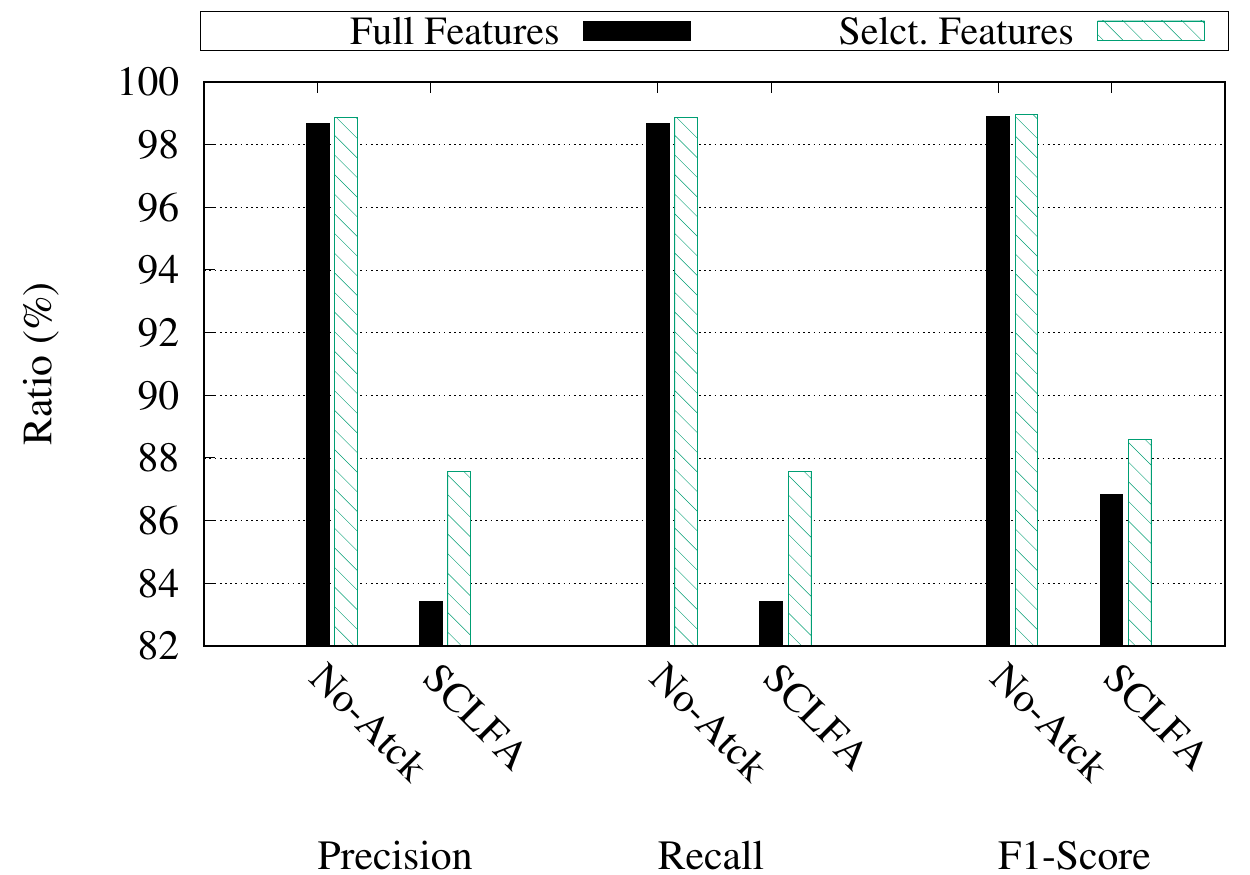}
			\caption{\small Drebin-API}
			\label{fig:fig11a}
 	\end{subfigure} 
   \begin{subfigure}{0.32\textwidth}
 		\centering
 		\includegraphics[width=1.01\linewidth]{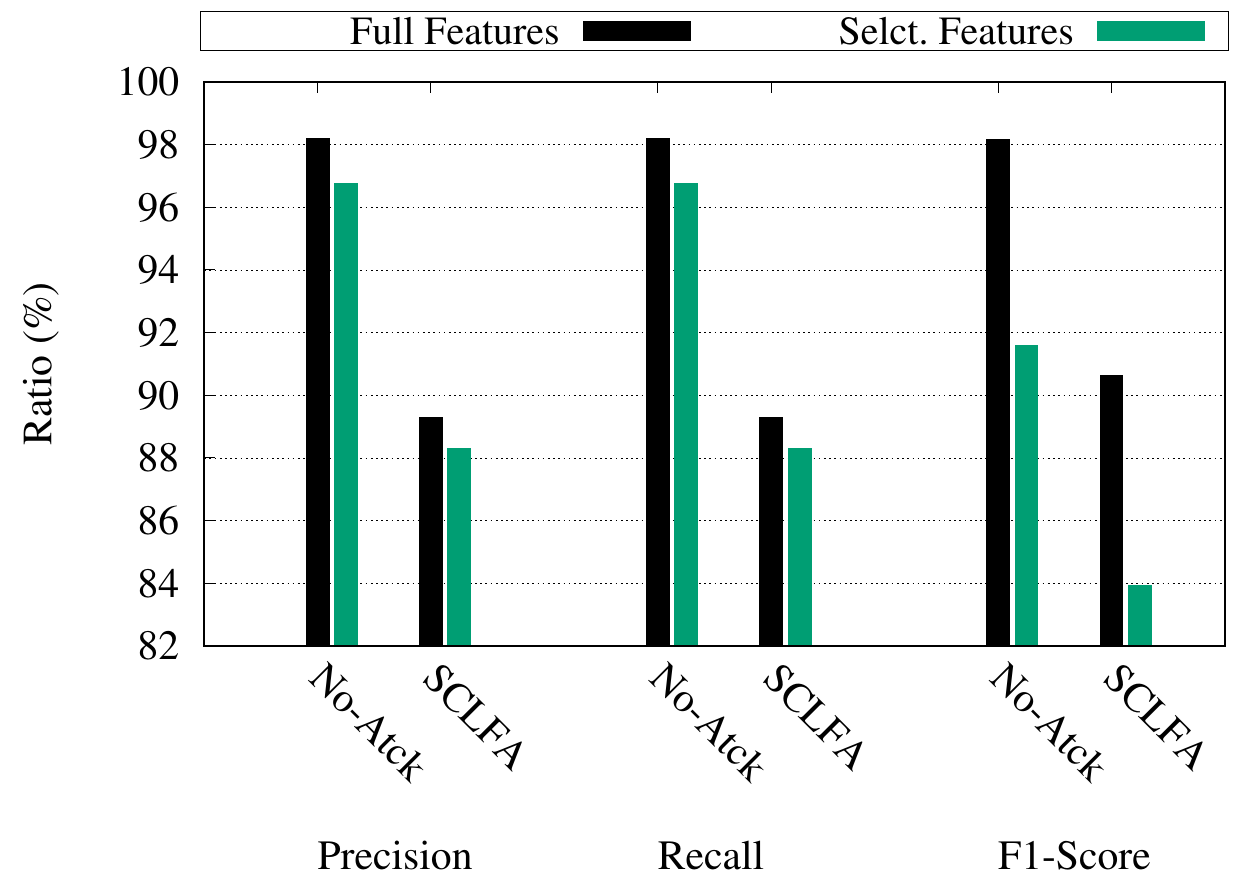}
			\caption{\small Drebin-permission}
			\label{fig:fig11b}
 	\end{subfigure} 
    \begin{subfigure}{0.32\textwidth}
 		\centering 
 		\includegraphics[width=1.01\linewidth]{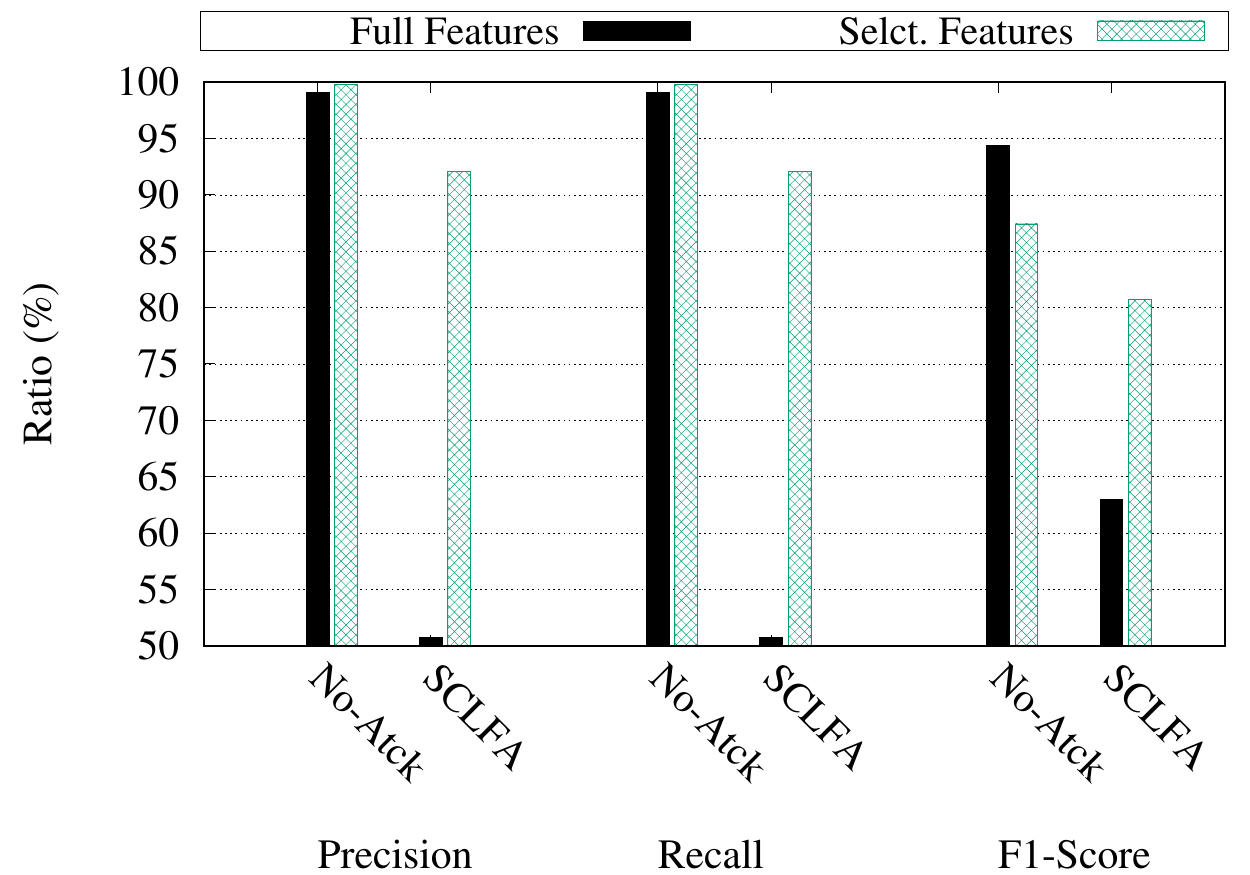}
		\caption{\small Drebin-intent}
			\label{fig:fig11c}
 	\end{subfigure}
 		\hfill
 	\begin{subfigure}{0.32\textwidth}
 		\centering
 		\includegraphics[width=1.01\linewidth]{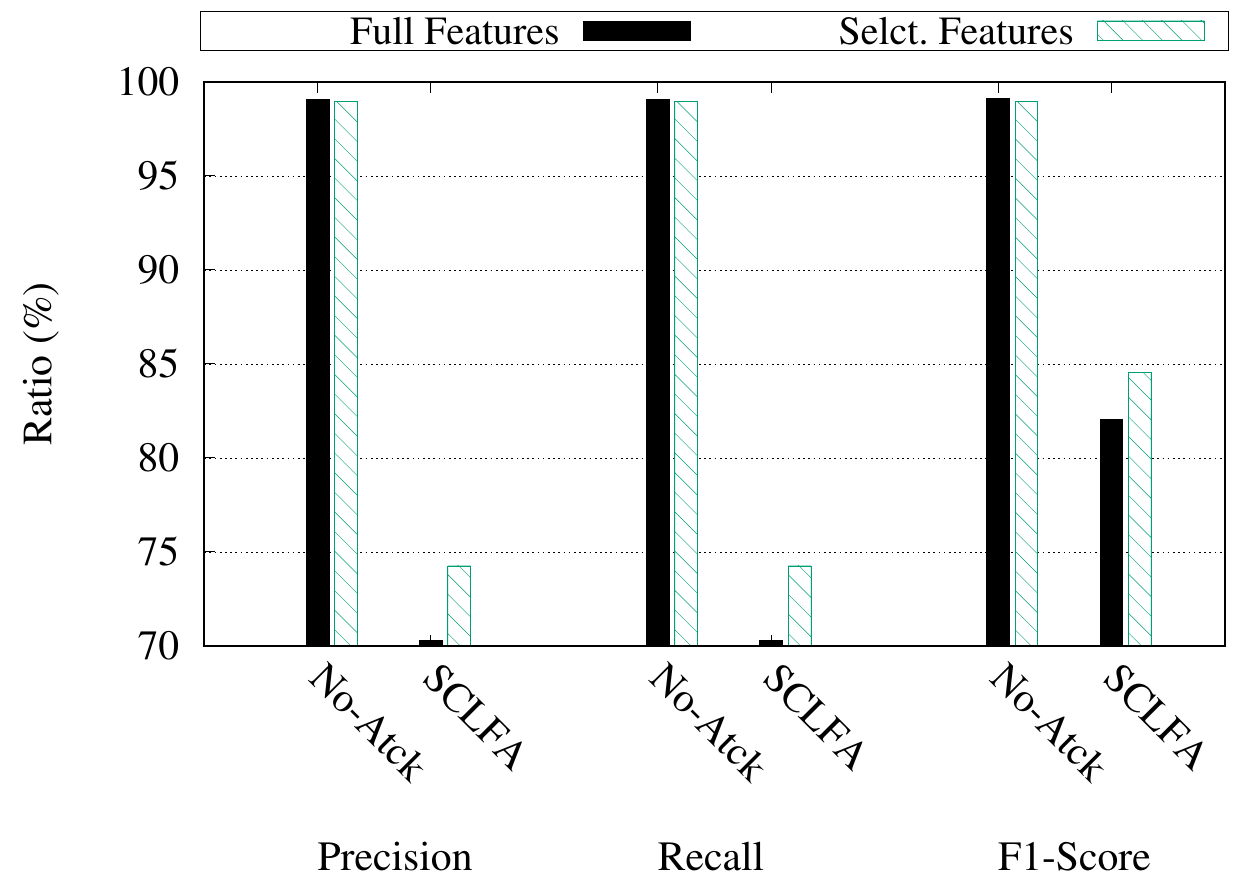}
			\caption{\small Contagio-API}
			\label{fig:fig11d}
 	\end{subfigure}  
    \begin{subfigure}{0.32\textwidth}
 		\centering 
 		\includegraphics[width=1.01\linewidth]{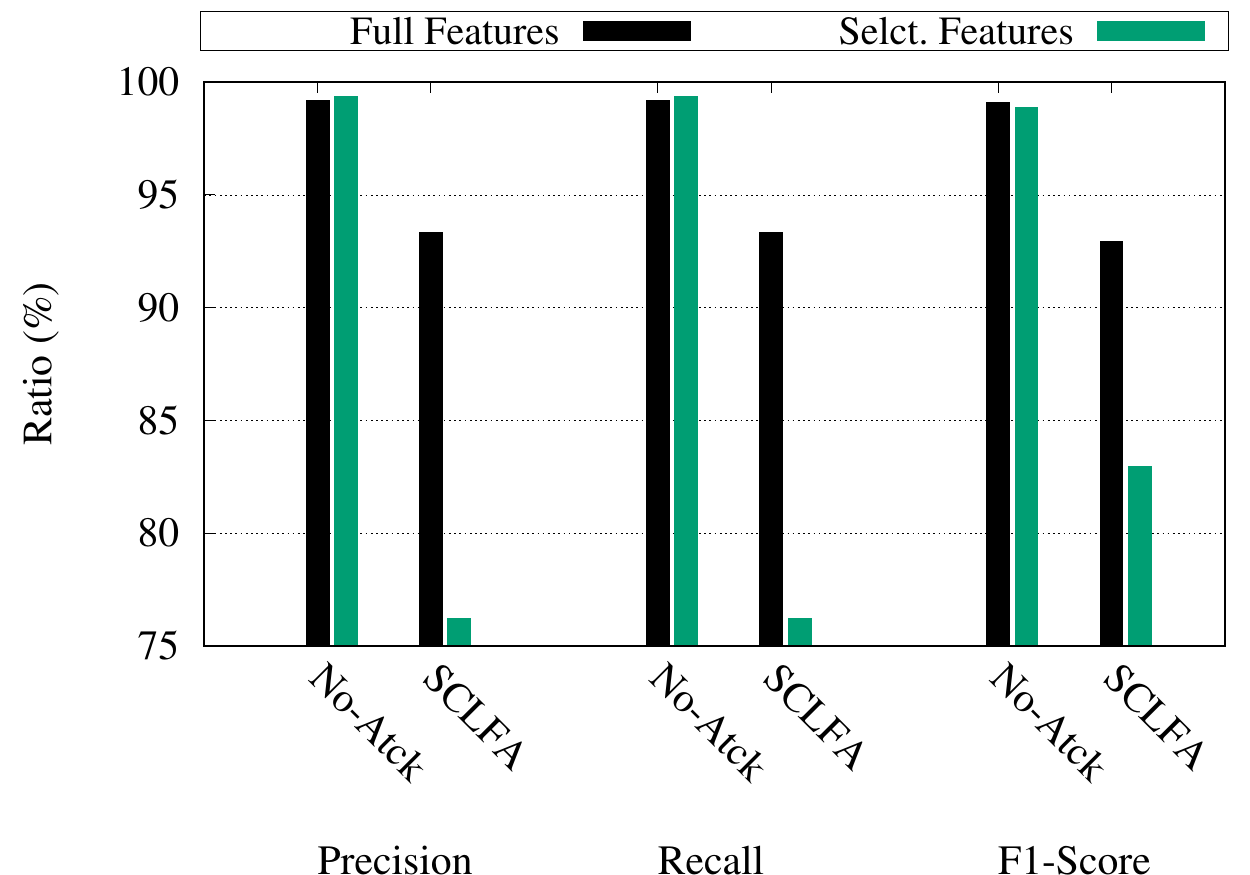}
			\caption{\small Contagio-permission}
			\label{fig:fig11e}
 	\end{subfigure}
   \begin{subfigure}{0.32\textwidth}
 		\centering
 		\includegraphics[width=1.01\linewidth]{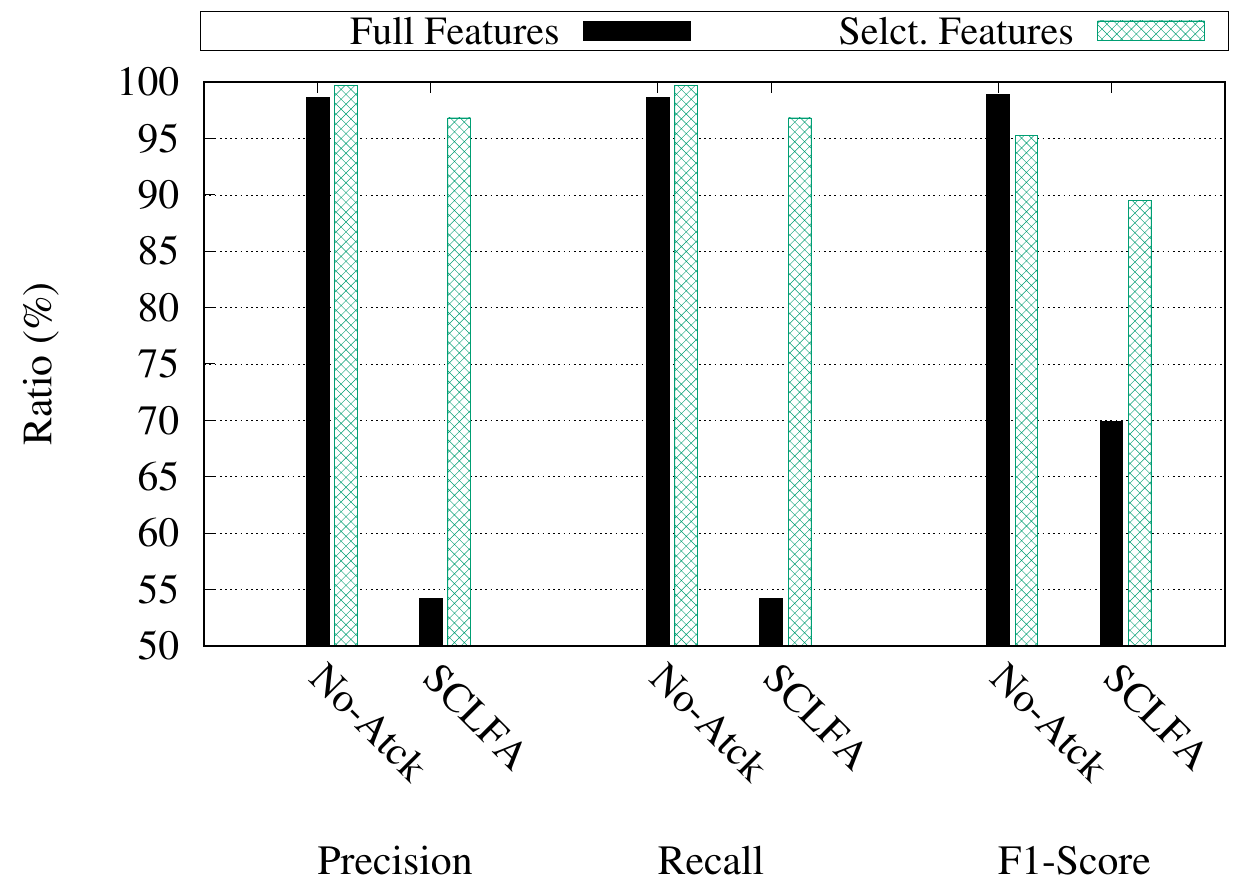}
		\caption{\small Contagio-intent}
			\label{fig:fig11f}
 	\end{subfigure} 
 	\begin{subfigure}{0.32\textwidth}
 		\centering
 		\includegraphics[width=1.01\linewidth]{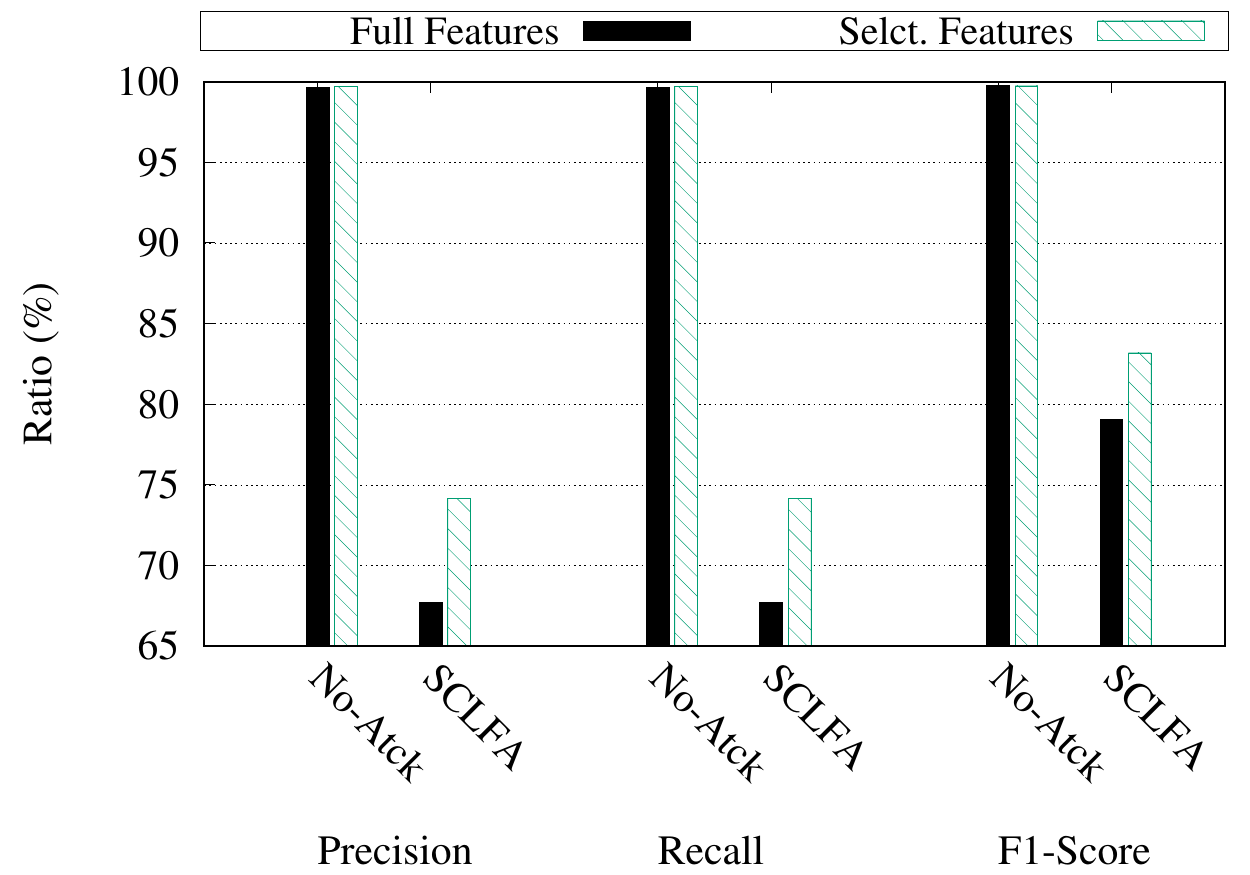}
			\caption{\small Genome-API}
			\label{fig:fig11g}
 	\end{subfigure}  
    \begin{subfigure}{0.32\textwidth}
 		\centering 
 		\includegraphics[width=1.01\linewidth]{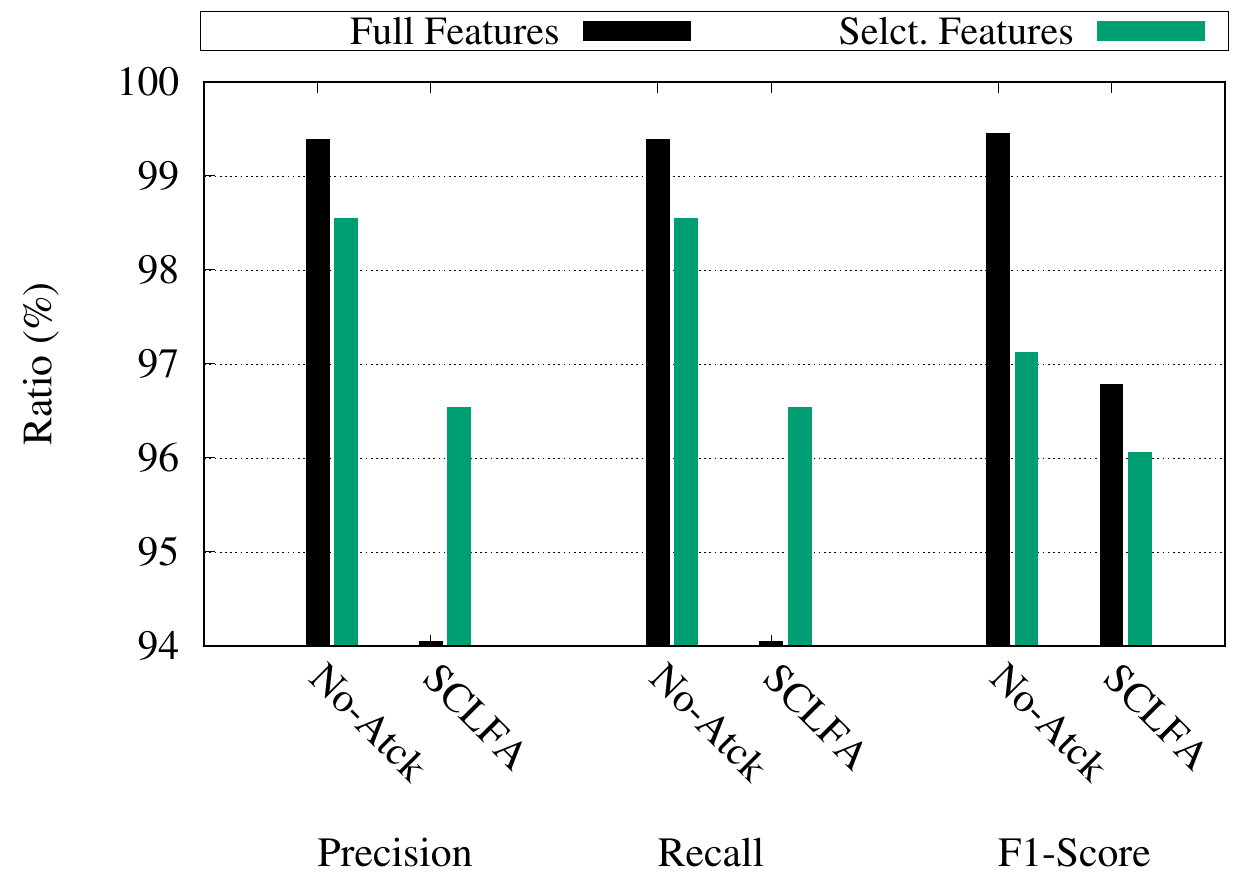}
			\caption{\small Genome-permission}
			\label{fig:fig11h}
 	\end{subfigure}
   \begin{subfigure}{0.32\textwidth}
 		\centering
 		\includegraphics[width=1.01\linewidth]{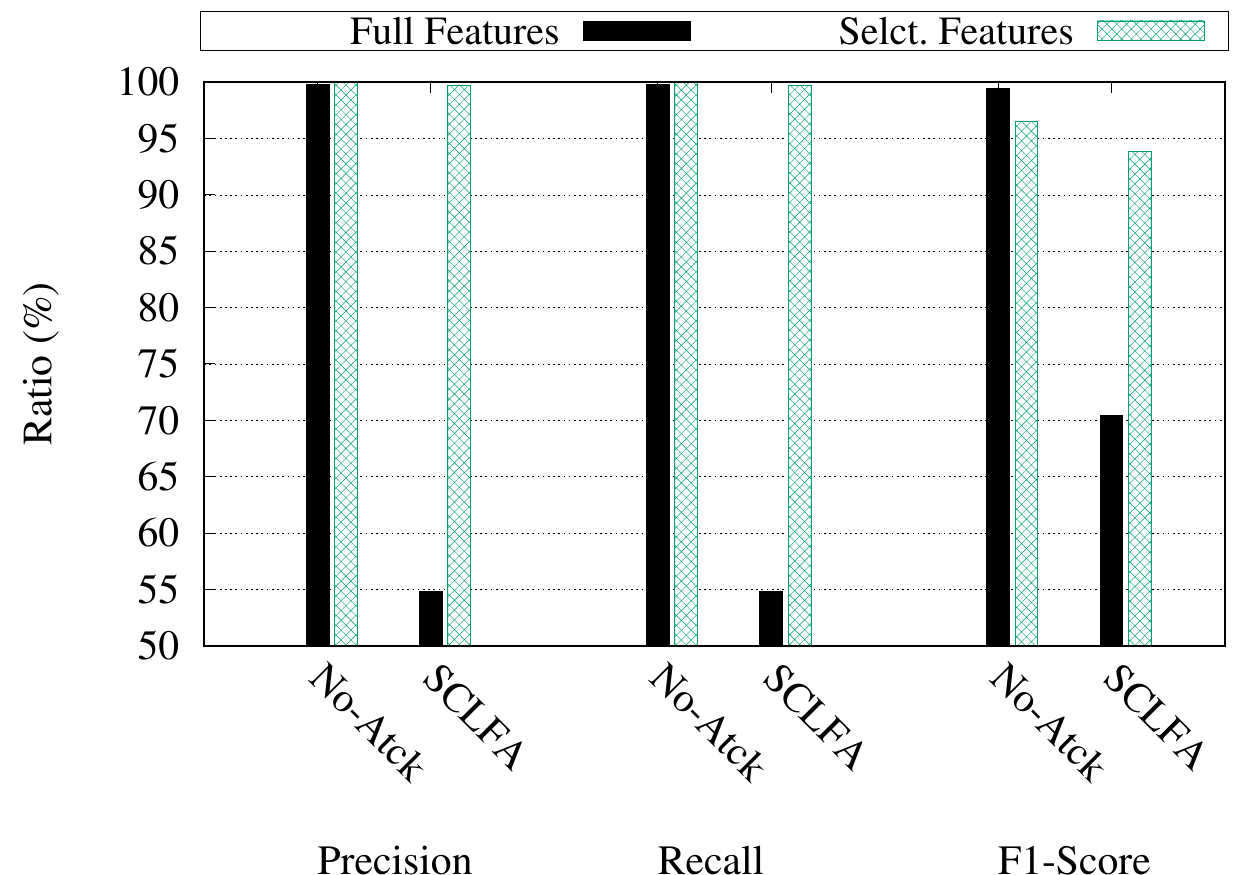}
		\caption{\small Genome-intent}
			\label{fig:fig11i}
 	\end{subfigure} 
 	\hfill
        \caption{\small Comparison between ATTACK algorithms with reference to \textcolor{black}{Precision, Recall, and F1-Score} for API, Intent and Permission features in various datasets.\vspace{-10px}}
			\label{fig:fig11}
		\end{figure*}

\textcolor{black}{In this test scenario, we aim to compare the defense algorithms (see Fig.~\ref{fig:fig10}) and attack method compares with the data without triggering the data, i.e., no-attack (see Fig.~\ref{fig:fig11}). Specifically, in Fig.~\ref{fig:fig10}, we provide Precision, Recall, and F1-score values for the different defense algorithms. Both Recall (Sensitivity) and Precision (Specificity) metrics indicate generated errors. The Recall is a measure that could show the rate of total detected malware. That is, the proportion of those correctly identified is the sum of all malware (i.e., those that are correctly identified by the malware plus those that are incorrectly detected by benign). Our goal in this section is to design a model with high Recall that is more appropriate to identify malware. To give more insight,   Figs.~\ref{fig:fig10a}-~\ref{fig:fig10c}, Figs.~\ref{fig:fig10d}-~\ref{fig:fig10f}, and Figs.~\ref{fig:fig10g}-~\ref{fig:fig10i} report the Permission, API and Intent data for the Drebin, Contagio, and Genome datasets, respectively. \textit{Three} considerations hold in this figures. \textit{i)} the value of Precision/Recall and even F1-score for KSSD algorithm and GAN-based algorithm are clearly lower than our LSD and CSD methods (as expected) and it confirms that our proposed defense algorithm is able to identify the more benign samples compared to other methods correctly. \textit{ii)} CSD algorithm have higher Precision and Recall values compared to LSD algorithm. \textit{iii)} in this feature group, our proposed algorithms have a higher Precision/Recall/F1-score value for Intent type features in all datasets compared to two other feature sets in which confirms our defense algorithm can detect more benign samples correctly in different data samples.}

Focusing on the attack consequences, Fig.~\ref{fig:fig11} presents the \textcolor{black}{Precision, Recall and F1-score} values for our attack algorithm, SCLFA, and the similar data when there is no any attack triggered for the ranked selected features and all three datasets with full features. In this figure, we understand that our attack strategy can completely fool the ML model and impose to falsify classification and exponentially decreases the \textcolor{black}{Precision and Recall} value. It can be seen that the diminishing rate is about 45\% for Intent features in all three datasets and its ratio is higher for Drebin dataset (see the \textcolor{black}{Precision/Recall} pair bars in Fig.~\ref{fig:fig11c}). It is considering that the feature selection has a positive influence on the ML performance. From the attacker point of view, it is essential to impact negatively the  ML model classification ability. Thus, our attack strategy gives the lead for such cases, and its adverse effects are more apparent in Contagio API features, where SCLFA influences on both selected features and full feature scenarios and can misclassify about 23\% of the samples (see the pair bars in Fig.~\ref{fig:fig11d}).

\subsubsection{Comparing methods based on FPR and Accuracy}\label{sec:5.b.2}
\begin{figure*}
    \centering
	\begin{subfigure}{0.45\textwidth}
 		\centering 
 		\includegraphics[width=1\linewidth]{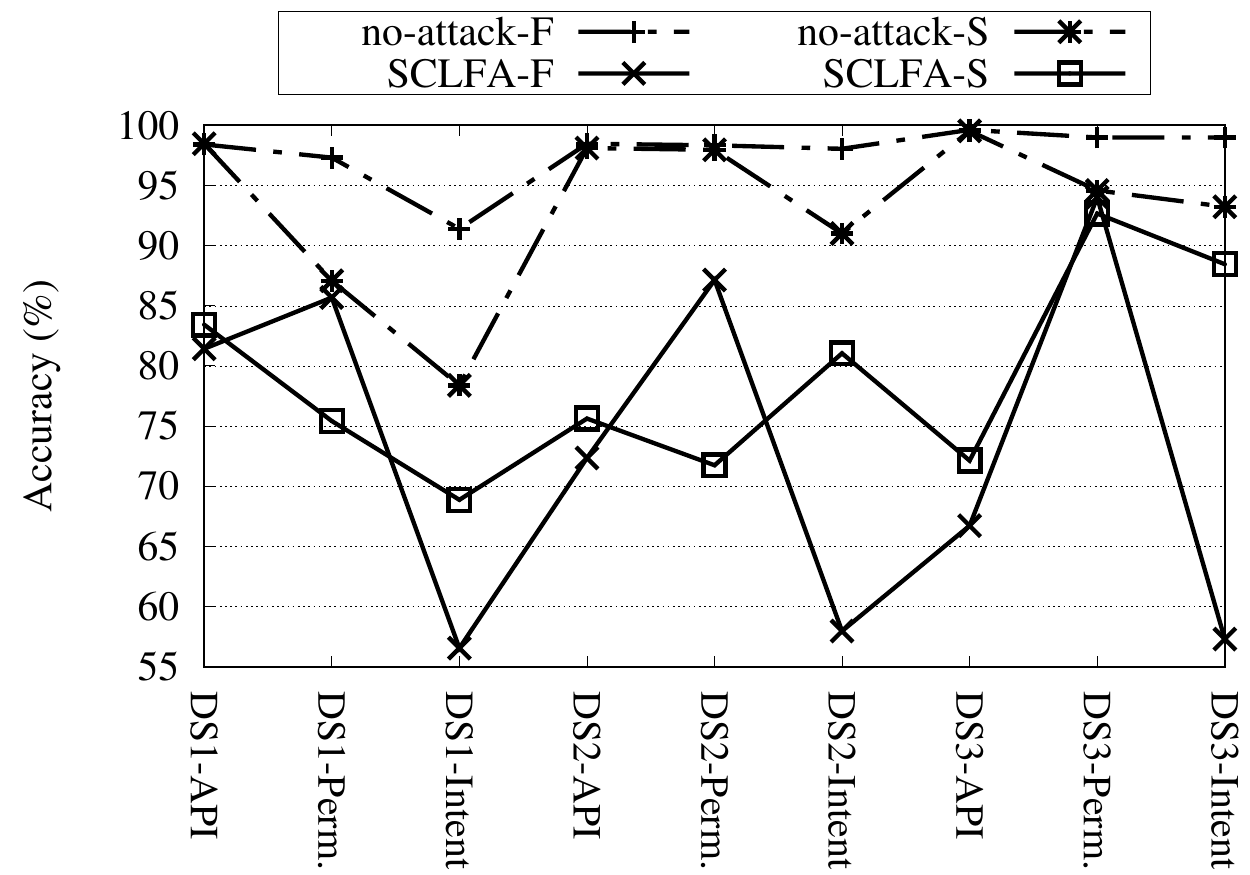}
			\caption{\small Accuracy Attack}
			\label{fig:fig5a}
 	\end{subfigure} 
   \begin{subfigure}{0.45\textwidth}
 		\centering
 		\includegraphics[width=1\linewidth]{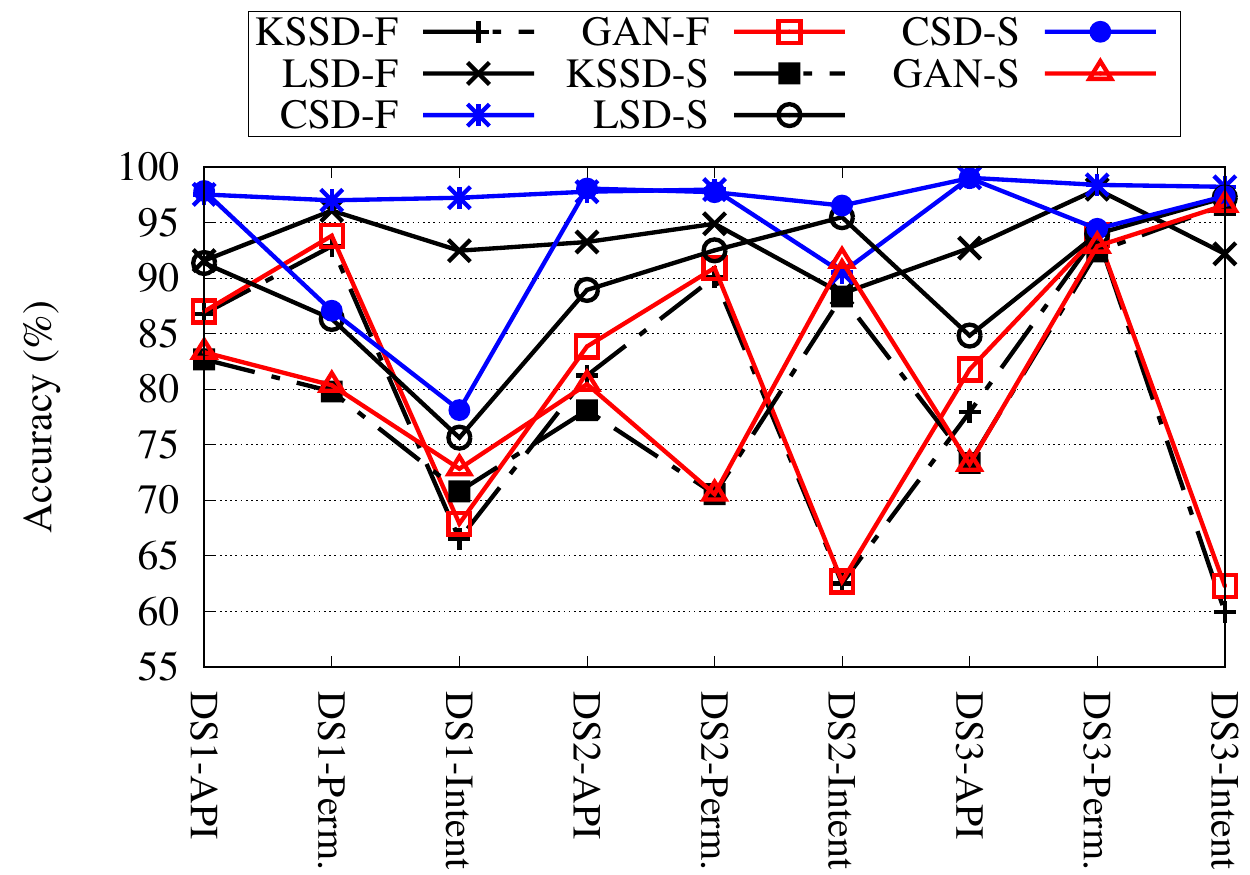}
			\caption{\small Accuracy Defense}
			\label{fig:fig5b}
 	\end{subfigure} 
 	\hfill
    \begin{subfigure}{0.45\textwidth}
 		\centering 
 		\includegraphics[width=1\linewidth]{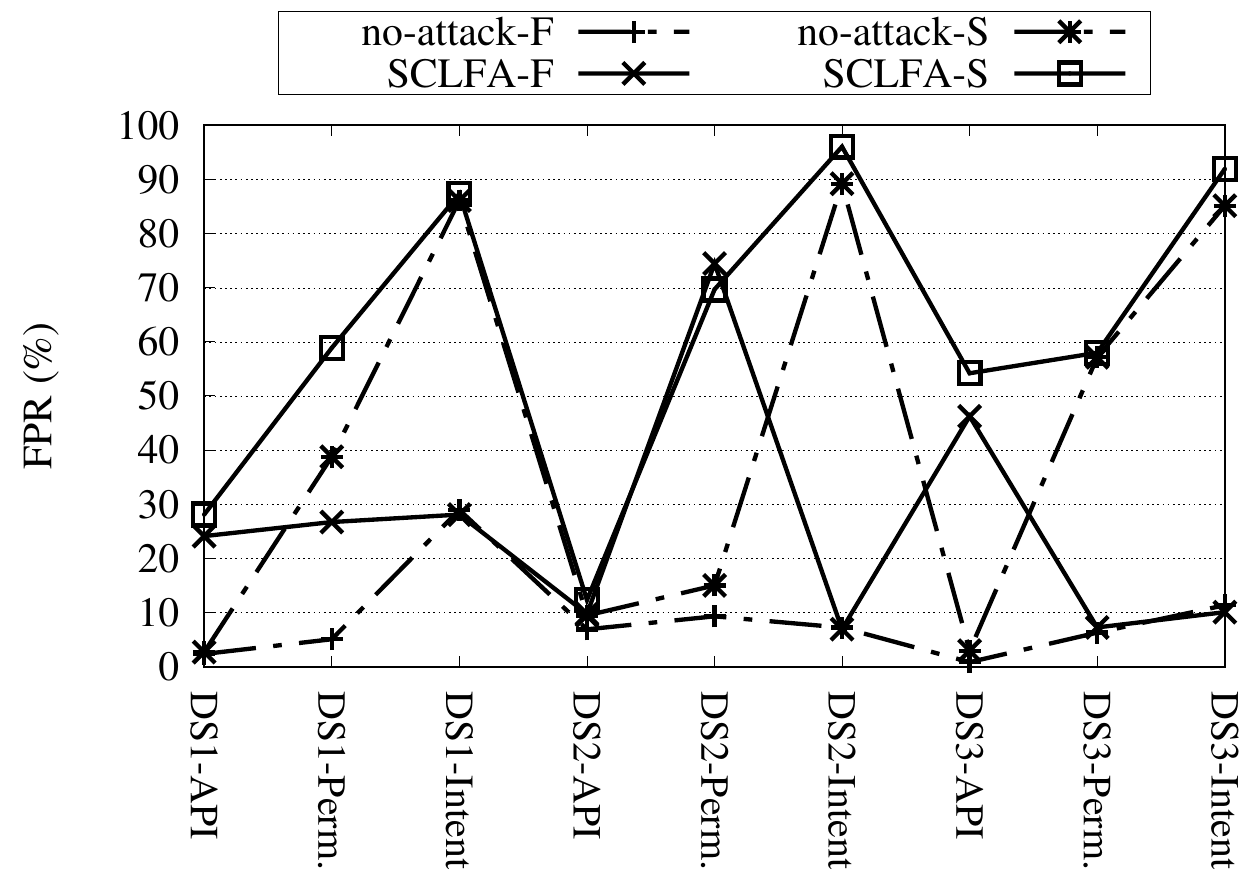}
		\caption{\small FPR Attack}
			\label{fig:fig5c}
 	\end{subfigure}
 	\begin{subfigure}{0.45\textwidth}
 		\centering
 		\includegraphics[width=1\linewidth]{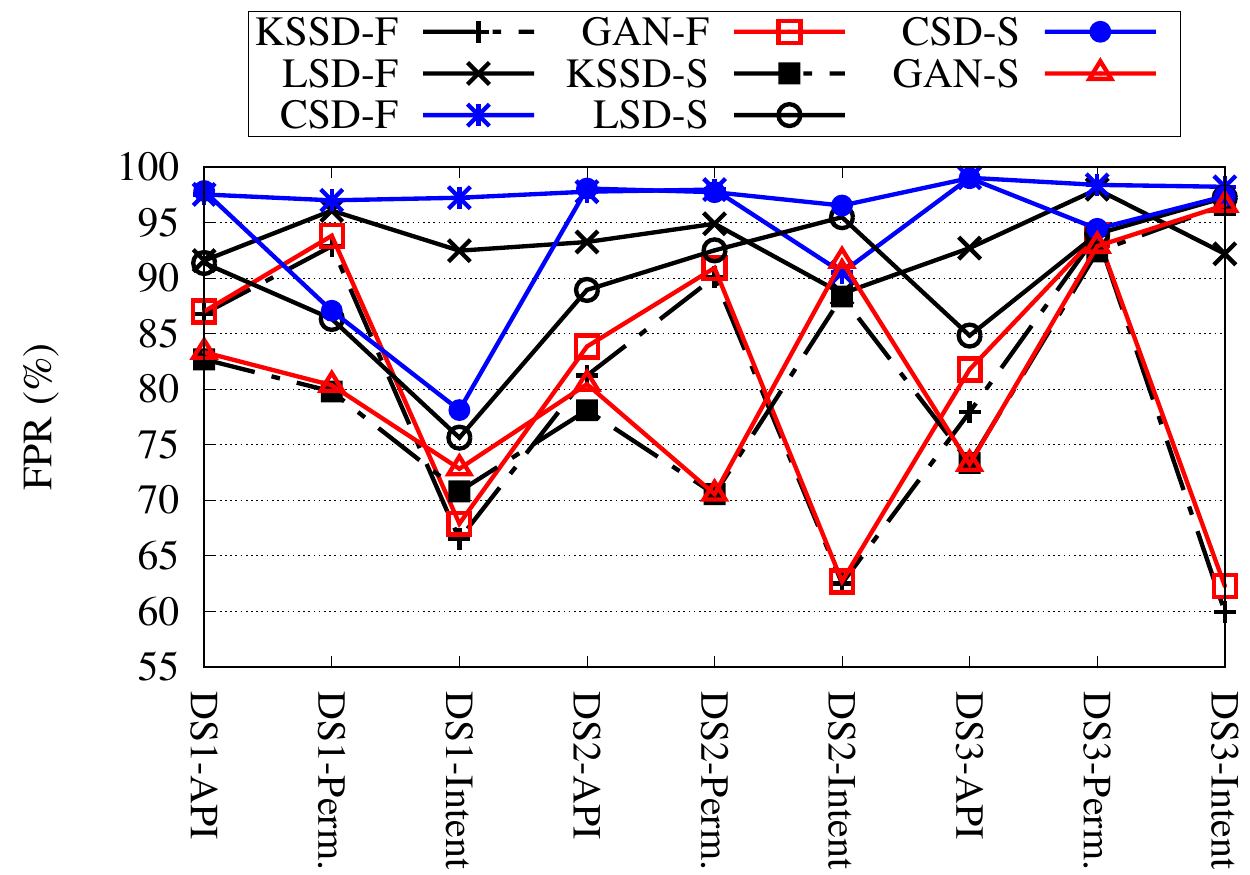}
			\caption{\small FPR Defense}
			\label{fig:fig5d}
 	\end{subfigure}  
        \caption{\small Comparison between attack and defense algorithms with reference to \textcolor{black}{Accuracy} and FPR for API, Intent and Permission features in various datasets.(DS1=Drebin; DS2=Contagio; DS3=Genome) and -F= full feature ; -S= selected features.\vspace{-10px}}
			\label{fig:fig5}
		\end{figure*}

In this part, we present the FPR and \textcolor{black}{Accuracy} values for the attack and defense algorithms for nine states, which consists of three features and three datasets and show them in Fig.~\ref{fig:fig5}. The attack algorithm aims to increase the FPR rate, and defense algorithm seeks to improve the \textcolor{black}{Accuracy} and decrease the FPR rate accordingly. Considering these points, we evaluate the algorithms on two mentioned scenarios: considering full features (labeled "-F") and feature selection scenario (labeled as "-Selc"). 

Focusing on Fig.~\ref{fig:fig5a}, we compare our attack algorithm (SCLFA) with the no attack mode for two mentioned scenarios. As we can understand from this figure, our SCLFA label flipping method compared to the no attack mode pose problem for all 9 states (i.e., listed in the x-axis of the figure) and results in lower \textcolor{black}{Accuracy} to all feature types and their major drastic are higher in Intent features for all datasets (see the `X' shape marks in lower part of Fig.~\ref{fig:fig5a}). In this case, the \textcolor{black}{Accuracy} has dropped more than 20\% compared to the absence of an attack. It confirms that the proposed attack method is more successful in attack to the Intents features. In all datasets, the algorithms behave roughly the same, and the reduction of the \textcolor{black}{Accuracy} of the API is more intense. As we see, the \textcolor{black}{Accuracy} of the proposed attack method in the case of using all data is lower than that of the 300 features.

Focusing on Fig.~\ref{fig:fig5b}, we compare the \textcolor{black}{Accuracy} of defense algorithms for full-featured and selected features scenarios. In this figure, almost in all cases, the CSD method can provide higher \textcolor{black}{Accuracy} than other defense algorithms (see the blue mark points in Fig.~\ref{fig:fig5b}) and is fallen in a range of (62\%,98\%). Therefore, it can detect more benign samples. The CSD method is more accurate than LSD in all 9 states, and its average \textcolor{black}{Accuracy} is about 95\%, 97.6\%, and 98.5\% for full feature consideration scenario in Drebin, Contagio and Gnome datasets, respectively while this value for KSSD defense algorithm is about 80\%, 79\%,and 77\%, respectively. 

Focusing on Fig.~\ref{fig:fig5c}, the FPR value of our attack algorithm (SCLFA) is compared to the time we have no attack in datasets. Concerning the Intent features in all datasets, the SCLFA algorithm, it has an FPR value and can fool more malware samples compared to other features in all datasets. In other words, increasing the FPR values means increasing the number of false positives, which is the goal of an attacker, and as can be seen from the comparison of Fig.~\ref{fig:fig5c} with Fig.~\ref{fig:fig5a}, by increasing FPR, the \textcolor{black}{Accuracy} value decreases. As a result, the results of these two figures confirm each other. 

Focusing on Fig.~\ref{fig:fig5d}, we compare the FPR values of defense algorithms. From this figure, we can understand that the CSD algorithm tries to decrease the FPR values more than two other defense algorithms in most of the states. It is important to note that having high \textcolor{black}{Accuracy} does not mean that the defense algorithm can successfully protect the dataset against the poisoning data, and it is essential to decrease the FPR in that state. Hence, we need to point-to-point check each state of this figure with a similar state in Fig.~\ref{fig:fig5b}. From these comparisons, we conclude that the CSD algorithm performs better than LSD and KSSD.  

\subsubsection{Comparing methods based on FNR and AUC}\label{sec:5.b.3}

In this part of our work, we compare the AUC and FNR values for attack and defense algorithms over three datasets using three different feature sets. In Table~\ref{tab3}, we present these results for two scenarios: with feature selection (WFS) and full features or without feature selection (WoFS). Concerning the FNR concept, it shows the misclassification rate of data in a dataset. In the flipping attack, the FNR rate increases, and it decreases with defense algorithms. Also, the AUC and FNR values indicate that by performing a flipping attack on the labels, FNR values increase and AUC values decrease. The AUC and FNR values is based on the equations \eqref{eq:eq20} and \eqref{eq:eq21} which relates to True Negative ($\Lambda$) and True positive ($\Omega$). FNR and AUC values with increasing FPR rates during an attack increase the $\Omega$ value and is pleasant for the attacker. Defense strategies try to decrease the FNR or miss rate of malware samples corrections and help to increase the AUC. In Table~\ref{tab3}, we understand that both of our defense algorithms have higher AUC and lower FNR rate for all datasets, and they confirm our recently mentioned points. 
\begin{table*}[!htpb]
\centering
\textcolor{black}{
\caption{\textcolor{black}{\small AUC and FNR Comparisons in percent ($\%$) for presented algorithms in various features, datasets in two test scenarios: WFS and WoFS. WFS= with feature selection; WoFS= without feature selection.}}
\label{tab3}
\footnotesize{
\begin{tabular}{|>{\centering\arraybackslash}m{0.02cm}|m{1.5cm}|p{1.4cm}|p{0.5cm}|p{0.5cm}|p{0.5cm}|p{0.5cm}||p{0.5cm}|p{0.5cm}|p{0.5cm}|p{0.5cm}||p{0.5cm}|p{0.5cm}|p{0.5cm}|p{0.5cm}|}
\hline
\rowcolor{white} 
\multicolumn{3}{|c|}{\textbf{Other ML Metrics}}&\multicolumn{12}{|c|}{\textbf{Datasets}}\\ \hline
\rowcolor{white} 
\multicolumn{3}{|c|}{\textbf{Ratio ($\%$)}}&\multicolumn{4}{|c|}{\textbf{Drebin}}&\multicolumn{4}{|c|}{\textbf{Contagio}}&\multicolumn{4}{|c|}{\textbf{Genome}}\\ \hline
\rowcolor{white}
\multicolumn{2}{|c|}{\textbf{Algorithms}}&\textbf{File Type}&\multicolumn{2}{|c|}{\textbf{WoFS}}&\multicolumn{2}{|c|}{\textbf{WFS}}&\multicolumn{2}{|c|}{\textbf{WoFS}}&\multicolumn{2}{|c|}{\textbf{WFS}}&\multicolumn{2}{|c|}{\textbf{WoFS}}&\multicolumn{2}{|c|}{\textbf{WFS}}\\\hline
\rowcolor{white}
\multicolumn{2}{|c|}{\textbf{}}&& {\textbf{FNR}}& {\textbf{AUC}}& {\textbf{FNR}}& {\textbf{AUC}}& {\textbf{FNR}}& {\textbf{AUC}}& {\textbf{FNR}}& {\textbf{FNR}}& {\textbf{FNR}}& {\textbf{AUC}}& {\textbf{FNR}}& {\textbf{AUC}}\\\hline\hline
\cellcolor[HTML]{dbfbff}&\cellcolor[HTML]{ffffa6}&\cellcolor[HTML]{fcf4eb}\textbf{Permission}&10.71&82.64&11.71&60.56&6.67&59.06&23.79&60.76&5.95&96.21&3.46&68.80\\\cline{3-15}
\cellcolor[HTML]{dbfbff}&\cellcolor[HTML]{ffffa6}&\cellcolor[HTML]{fcf4eb}\textbf{API}&16.59&83.22&12.43&80.78&29.70&94.35&25.75&92.91&32.29&74.40&25.86&70.24\\\cline{3-15}
\cellcolor[HTML]{dbfbff}\multirow{-3}{*}{\begin{sideways}{\textbf{Attack}}\end{sideways}}&\cellcolor[HTML]{ffffa6}\multirow{-3}{*}{\textbf{SCLFA}}&\cellcolor[HTML]{fcf4eb}\textbf{Intents}&49.21&62.97&7.94&42.28&45.83&95.87&3.21&43.55&45.22&94.26&0.31&48.32\\\hline\hline
\cellcolor[HTML]{dbfbff}&\cellcolor[HTML]{ffffa6}&\cellcolor[HTML]{fcf4eb}\textbf{Permission}&2.17&93.90&2.33&70.77&1.65&80.32&0.81&61.64&1.29&93.89&2.05&68.65\\\cline{3-15}
\cellcolor[HTML]{dbfbff}&\cellcolor[HTML]{ffffa6}&\cellcolor[HTML]{fcf4eb}\textbf{API}&6.62&90.77&5.65&88.44&6.59&95.34&11.13&94.03&7.40&96.59&12.15&55.23\\\cline{3-15}
\cellcolor[HTML]{dbfbff}&\cellcolor[HTML]{ffffa6}\multirow{-3}{*}{\textbf{LSD}}&\cellcolor[HTML]{fcf4eb}\textbf{Intents}&10.20&91.97&0.43&43.70&6.28&89.42&0.75&50.07&7.27&91.82&0.53&50.41\\\cline{2-15}
\cellcolor[HTML]{dbfbff}&\cellcolor[HTML]{ffffa6}&\cellcolor[HTML]{fcf4eb}\textbf{Permission}&1.44&95.08&3.22&73.98&0.94&93.18&0.84&91.36&1.33&97.06&1.58&69.07\\\cline{3-15}
\cellcolor[HTML]{dbfbff}&\cellcolor[HTML]{ffffa6}&\cellcolor[HTML]{fcf4eb}\textbf{API}&2.53&98.40&2.04&98.28&1.47&95.05&0.42&91.39&0.84&98.42&0.65&96.83\\\cline{3-15}
\cellcolor[HTML]{dbfbff}&\cellcolor[HTML]{ffffa6}\multirow{-3}{*}{\textbf{CSD}}&\cellcolor[HTML]{fcf4eb}\textbf{Intents}&2&93.74&0.54&45.60&1.75&93.18&0.55&50.31&1.10&94.12&0.40&51.43\\\cline{2-15}
\cellcolor[HTML]{dbfbff}&\cellcolor[HTML]{ffffa6}&\cellcolor[HTML]{fcf4eb}\textbf{Permission}&48.95&47.00&11.06&70.93&47.91&0.54&25.28&61.19&33.62&55.49&3.51&69.14\\\cline{3-15}
\cellcolor[HTML]{dbfbff}\multirow{-4}{*}{\begin{sideways}{\textbf{Defenses}}\end{sideways}}&\cellcolor[HTML]{ffffa6}&\cellcolor[HTML]{fcf4eb}\textbf{API}&11.11&87.37&12.46&78.92&16.96&93.80&20.78&95.58&27.26&59.77&24.78&70.76\\\cline{3-15}
\cellcolor[HTML]{dbfbff}&\cellcolor[HTML]{ffffa6}\multirow{-3}{*}{\textbf{GANX~\cite{taheri2019can}}}&\cellcolor[HTML]{fcf4eb}\textbf{Intents}&87.35&8.96&6.99&43.50&33.46&57.75&2.10&71.60&26.36&60.84&0.57&83.29\\\cline{2-15}
\cellcolor[HTML]{dbfbff}&\cellcolor[HTML]{ffffa6}&\cellcolor[HTML]{fcf4eb}\textbf{Permission}&4.95&91.17&11.58&70.59&3.53&61.88&25.19&90.94&5.60&90.91&3.72&68.57\\\cline{3-15}
\cellcolor[HTML]{dbfbff}&\cellcolor[HTML]{ffffa6}&\cellcolor[HTML]{fcf4eb}\textbf{API}&10.78&86.27&12.46&78.92&19.81&94.45&23.27&94.25&19.91&72.48&24.73&71\\\cline{3-15}
\cellcolor[HTML]{dbfbff}&\cellcolor[HTML]{ffffa6}\multirow{-3}{*}{\textbf{KSSD~\cite{paudice2018label}}}&\cellcolor[HTML]{fcf4eb}\textbf{Intents}&39.10&70.93&6.91&42.39&35.11&89.02&2.35&44.58&41.79&90.04&0.30&50.51\\\hline
\end{tabular}}
}
\end{table*}

\subsubsection{Computational complexity comparisons}\label{sec:5.b.4}
    \textcolor{black}{In this part of the paper, we compare the computational complexity of our attack and the defense algorithms against (KSSD)~\cite{paudice2018label} and GAN-based Defense~\cite{taheri2019can}. Table~\ref{tab4} compares the time required for the testing phase of various datasets and different features among the different proposed algorithms for ranked features using RF and without feature selection methods. In this table, focusing on the defense algorithms, the implementation of KSSD defense is the fastest compared to LSD, CSD, and GAN-based algorithms. The reason behind it is that it randomly selects and modifies the label of the features. However, its \textcolor{black}{Accuracy} is much lower than LSD and CSD algorithms (see the curves in Fig.~\ref{fig:fig5b}). 
    However, both LSD and CSD algorithms require processing of the DL algorithm on the malicious dataset, with the LSD and GAN being the slowest method among the \textit{four} methods in correcting the labels of poisoning samples. }
    
    Focusing on the computational complexity of ranked features scenario, we can understand that the LSD and CSD methods, running faster in Intents and Permissions features and these results are even quicker when we compared them with API features cases. We can conclude that the distribution of API features may require more computations in calculating LSD and CSD, and this is a normal behavior of the algorithms. Because in the feature selection, we select 300 API features that have the highest rank based on the RF feature selection algorithm, and this covers about 95\% of the API features from every data sets, but the selection of 300 features from Intents and Permission intends selecting at most 20\% of the main features. Therefore, the computational complexity of the proposed algorithm over the API features when the 300 features are chosen is close to the state in which all the features are used. 
    
    Focusing on the computational complexity of full features comparisons scenario, we can see that since the number of API features is much less than the Intents and Permissions, so the computational time of proposed algorithms is less on these features. Similarly, we can conceive the same results for the computational complexity of proposed algorithms on the Permission features than the results of Intents features (see the Permission row values for WFoS cases in all datasets in Table~\ref{tab4}). Also, from Table~\ref{tab4} we can realize that the proposed methods are slower than the KSSD method. However, as we understand from the comparisons of ML metrics, the KSSD method is a weak method using for label flipping attack (LFA) compared to our proposed defense algorithms.

Additionally, the computational time of CSD with taking into account the high \textcolor{black}{Accuracy} of this method and its take less running time compared to the LSD method. So, this behavior converts the CSD method into an attractive way to defend against the LFA. Another point to be added about the time of the LSD method is to consider the structure of the method in which a CNN network is used, whose time complexity is at least $\mathcal{O}(n ^ 3)$ and this can be the major drawback when it compares to the CSD algorithm.


\begin{table*}[!htpb]
\centering
\textcolor{black}{
\caption{\textcolor{black}{\small Computational complexity comparisons in seconds ($s$) for presented algorithms in various features, datasets in two test scenarios: WFS and WoFS. WFS= with feature selection; WoFS= without feature selection.}}
\label{tab4}
\footnotesize{
\begin{tabular}{|>{\centering\arraybackslash}m{0.1cm}|m{1.5cm}|p{1.6cm}|p{0.55cm}|p{0.55cm}||p{0.55cm}|p{0.55cm}||p{0.55cm}|p{0.55cm}|}
\hline
\rowcolor{white} 
\multicolumn{3}{|c|}{\textbf{Computational Complexity}}&\multicolumn{6}{|c|}{\textbf{Datasets}}\\ \hline
\rowcolor{white} 
\multicolumn{3}{|c|}{\textbf{Time (s)}}&\multicolumn{2}{|c|}{\textbf{Drebin}}&\multicolumn{2}{|c|}{\textbf{Contagio}}&\multicolumn{2}{|c|}{\textbf{Genome}}\\ \hline
\rowcolor{white}
\multicolumn{2}{|c|}{\textbf{Algorithms}}&\textbf{File Type}&\multicolumn{1}{|c|}{\textbf{WoFS}}&\multicolumn{1}{|c|}{\textbf{WFS}}&\multicolumn{1}{|c|}{\textbf{WoFS}}&\multicolumn{1}{|c|}{\textbf{WFS}}&\multicolumn{1}{|c|}{\textbf{WoFS}}&\multicolumn{1}{|c|}{\textbf{WFS}}\\\hline\hline
\cellcolor[HTML]{dbfbff}&\cellcolor[HTML]{ffffa6}&\cellcolor[HTML]{fcf4eb}\textbf{Permission}&140.09&4.04&87.66&3.56&130.10&3.11\\\cline{3-9}
\cellcolor[HTML]{dbfbff}&\cellcolor[HTML]{ffffa6}&\cellcolor[HTML]{fcf4eb}\textbf{API}&7.14&4.71&4.84&3.88&4.21&3.74\\\cline{3-9}
\cellcolor[HTML]{dbfbff}\multirow{-3}{*}{\begin{sideways}{\textbf{Attack}}\end{sideways}}&\cellcolor[HTML]{ffffa6}\multirow{-3}{*}{\textbf{SCLFA}}&\cellcolor[HTML]{fcf4eb}\textbf{Intents}&150.99&3.83&209.89&2.87&106.07&2.92\\\hline\hline
\cellcolor[HTML]{dbfbff}&\cellcolor[HTML]{ffffa6}&\cellcolor[HTML]{fcf4eb}\textbf{Permission}&385.79&101.16&417.62&107.81&348.62&106.02\\\cline{3-9}
\cellcolor[HTML]{dbfbff}&\cellcolor[HTML]{ffffa6}&\cellcolor[HTML]{fcf4eb}\textbf{API}&123.91&114.64&117.35&112.75&109.87&105.38\\\cline{3-9}
\cellcolor[HTML]{dbfbff}&\cellcolor[HTML]{ffffa6}\multirow{-3}{*}{\textbf{LSD}}&\cellcolor[HTML]{fcf4eb}\textbf{Intents}&963.97&105.17&747.98&96.81&501.85&108.10\\\cline{2-9}\cline{2-9}
\cellcolor[HTML]{dbfbff}&\cellcolor[HTML]{ffffa6}&\cellcolor[HTML]{fcf4eb}\textbf{Permission}&148.15&11.50&118.77&9.51&123.45&9.16\\\cline{3-9}
\cellcolor[HTML]{dbfbff}&\cellcolor[HTML]{ffffa6}&\cellcolor[HTML]{fcf4eb}\textbf{API}&21.76&15.77&17.22&13.27&14.56&12.77\\\cline{3-9}
\cellcolor[HTML]{dbfbff}\multirow{-4}{*}{\begin{sideways}{\textbf{Defenses}}\end{sideways}}&\cellcolor[HTML]{ffffa6}\multirow{-3}{*}{\textbf{CSD}}&\cellcolor[HTML]{fcf4eb}\textbf{Intents}&281.83&11.26&235.24&9.21&198.63&11.42\\\cline{2-9}\cline{2-9}
\cellcolor[HTML]{dbfbff}&\cellcolor[HTML]{ffffa6}&\cellcolor[HTML]{fcf4eb}\textbf{Permission}&95.90&5.20&83.91&4.15&90.83&5.16\\\cline{3-9}
\cellcolor[HTML]{dbfbff}&\cellcolor[HTML]{ffffa6}&\cellcolor[HTML]{fcf4eb}\textbf{API}&9.95&7.59&8.53&6.46&8.41&6.42\\\cline{3-9}
\cellcolor[HTML]{dbfbff}&\cellcolor[HTML]{ffffa6}\multirow{-3}{*}{\textbf{KSSD~\cite{paudice2018label}}}&\cellcolor[HTML]{fcf4eb}\textbf{Intents}&210.99&5.17&206.77&4.12&146.55&5.12\\\cline{2-9}
\cellcolor[HTML]{dbfbff}&\cellcolor[HTML]{ffffa6}&\cellcolor[HTML]{fcf4eb}\textbf{Permission}&425.13&211.64&471.33&194.55&394.65&176.38\\\cline{3-9}
\cellcolor[HTML]{dbfbff}&\cellcolor[HTML]{ffffa6}&\cellcolor[HTML]{fcf4eb}\textbf{API}&94.23&67.45&86.56&64.75&75.34&57.14\\\cline{3-9}
\cellcolor[HTML]{dbfbff}&\cellcolor[HTML]{ffffa6}\multirow{-3}{*}{\textbf{GAN~\cite{taheri2019can}}}&\cellcolor[HTML]{fcf4eb}\textbf{Intents}&515.41&276.54&495.32&209.21&436.97&196.45\\\hline
\end{tabular}}
}
\end{table*}

\section{Discussions}\label{discussion}
In the following, we explain the achievements and some constraints on our attack and defense algorithms. From the results, we can conclude that the proposed methods based on semi-supervised learning can modify the flipped labels to increase the \textcolor{black}{Accuracy} of classification methods, including CNN. Despite the promising results achieved by our attack and defense algorithms, it is clear that our approaches have some intrinsic limitations. Firstly, the critical point in the proposed methods is the need for more calculations. Notably, the use of a CNN in the LSD method increases the computational complexity of this method. Secondly, the proposed malware detection algorithms implementing static features, the features are binary and are in the sparse matrix. Hence, it is easier to calculate clustering measures. However, for other applications, it may not be possible to perform calculations of the clustering measures efficiently.

Another limitation of our defense methods is the classification algorithm used in this paper. Formally speaking, in this work, we design a three-layer CNN, which has high \textcolor{black}{Accuracy} and use CNN to investigate the results of the proposed algorithms. The classification \textcolor{black}{Accuracy} with other classification algorithms is another issue that needs to be addressed. The \textcolor{black}{Accuracy} and FPR in the comparison figures indicate that when the feature selection applied (second scenario), the proposed methods still have acceptable values for these two measurements, but LSD and CSD algorithms running faster than the case that testing of the full features is employed (as expected and visible).

\section{Conclusions and future work}\label{conclusion}

In this paper, we design an attack and two defense algorithms which target Android malware detection system, namely a Silhouette-based label flipping attack (SCLFA), a label-based semi-supervised defense algorithm (LSD), and a clustering-based semi-supervised defense (CSD) algorithm. We compare our defense algorithms against the  KNN-based label flipping attack on Android mobile dataset using \textit{three} public datasets, i.e., the Drebin, Genome, and Contagio datasets, using different API, Intent and Permission features. We test our models on a CNN classification algorithm. The comparison of proposed CSD and LSD methods against the KSSD method reveals that the proposed methods have higher \textcolor{black}{Accuracy} than the KSSD, while the KSSD algorithm is faster. To be precise, the CSD algorithm, although slightly slower than the KSSD algorithm, but since in many cases, it has approximately 19\% higher \textcolor{black}{Accuracy} than the KSSD and has about 15\% lower FPR compared to the KSSD. For future work, we suggest using semi-supervised methods based on deep learning techniques, such as autoencoder and various types of GAN networks. They can be used along with clustering techniques. Using these methods as an ensemble learning can provide excellent results against label flipping attacks.

\section*{Acknowledgment}
 Mauro Conti and Mohammad Shojafar are supported by a Marie Curie Fellowship funded by the European Commission (agreement PCIG11-GA-2012-321980) and (agreement MSCA-IF-GF-2019-839255), respectively.\vspace{-10px}
 \section*{Conflict of interest}
There is no any conflict of interest for the paper.
\Urlmuskip=0mu plus 1mu\relax
\bibliographystyle{spmpsci}
\bibliography{references.bib}
\clearpage
\section*{Biographies}\label{sec:11}
\vspace{-10px}
\noindent\begin{minipage}{0.14\textwidth}
\includegraphics[width=1.15in,height=1.15in,clip,keepaspectratio]{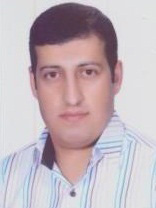} 
\end{minipage}%
\hfill%
\begin{minipage}{0.84\textwidth}
\textbf{Rahim~Taheri} received his B.Sc. degree of Computer engineering from Bahonar Technical College of Shiraz and M.Sc. degree of computer networks at the Shiraz University of Technology in 2007 and 2015, respectively. Now he is a Ph.D. candidate on Computer Networks at the Shiraz University of Technology. In February 2018, he joined to SPRITZ Security \& Privacy Research Group at the University of Padua as a visiting Ph.D. student. His main research interests include ML, data mining,  network securities and heuristic algorithms. He currently focused on adversarial machine and deep learning as a new trend in computer security.
\end{minipage}%

\hfill \break

\noindent\begin{minipage}{0.14\textwidth}
\includegraphics[width=1in,height=1.15in,clip,keepaspectratio]{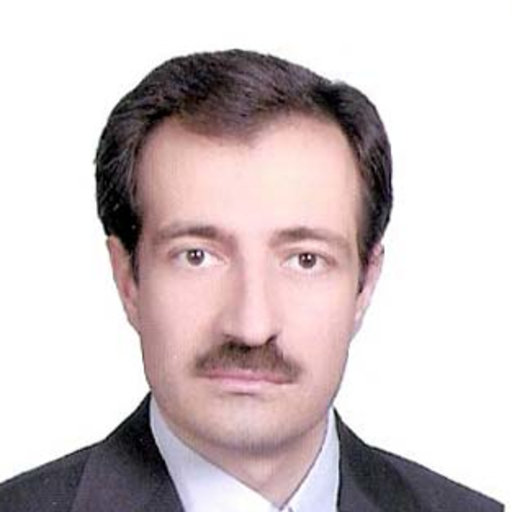}  
\end{minipage}%
\hfill%
\begin{minipage}{0.84\textwidth}
 \textbf{Reza~Javidan} received M.Sc. Degree in Computer Engineering (Machine Intelligence and Robotics) from Shiraz University in 1996. He received a Ph.D. degree in Computer Engineering (Artificial Intelligence) from Shiraz University in 2007. Dr. Javidan has many publications in international conferences and journals regarding Image Processing, Underwater Wireless Sensor Networks (UWSNs) and Software Defined Networks (SDNs). His major fields of interest are Network security, Underwater Wireless Sensor Networks (UWSNs), Software Defined Networks (SDNs), Internet of Things, artificial intelligence, image processing, and SONAR systems. Dr. Javidan is an associate professor in the Department of Computer Engineering and Information Technology at the Shiraz University of Technology. For additional information: \url{https://scholar.google.com/citations?user=7XhPDVEAAAAJ&hl=en}
 
 \end{minipage}%

\hfill \break

\noindent\begin{minipage}{0.14\textwidth}
\includegraphics[width=0.9in,height=1.15in,clip,keepaspectratio]{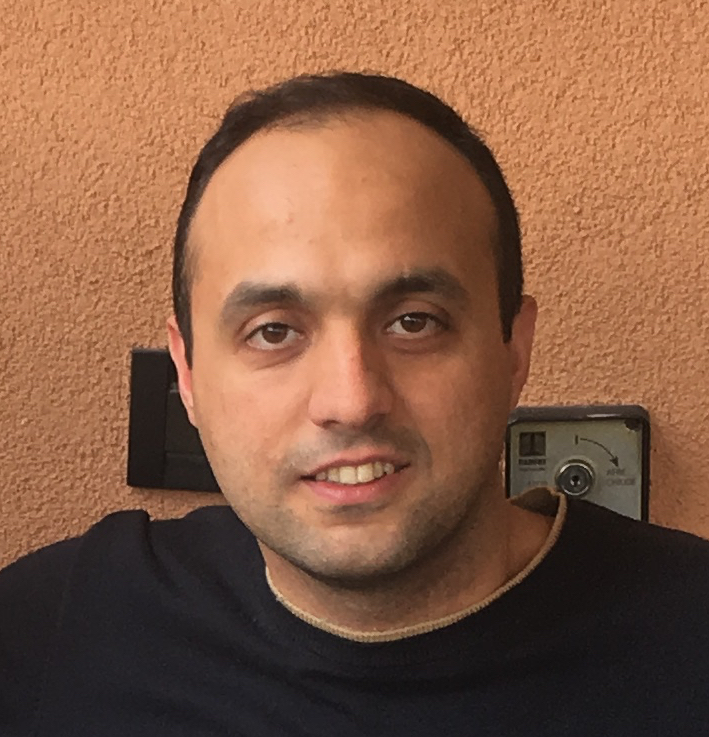}
\end{minipage}%
\hfill
\begin{minipage}{0.84\textwidth}	
\textbf{Mohammad Shojafar} is a Marie Curie Fellow, Intel Innovator, and Senior Researcher working on PRISENODE project in the SPRITZ Security and Privacy Research group at the University of Padua, Italy. Before starting MSCA project, he was a Senior Researcher working on network security project at the Ryerson University in 2019. Also, he was CNIT Senior Researcher at the University of Rome Tor Vergata contributed on European H2020 ``SUPERFLUIDITY'' project. Mohammad also was a principal investigator on an Italian SDN security and privacy (60,000 euro) supported by the University of Padua in 2018. He also was contributed to some Italian projects in telecommunications like GAUChO — A Green Adaptive Fog Computing and Networking Architecture (400,000 euro), S2C: Secure, Software-defined Cloud (30,000 euro), and SAMMClouds-Secure and Adaptive Management of Multi-Clouds (30,000 euro) collaborating among Italian universities. He received the Ph.D. degree from Sapienza University of Rome, Italy, in 2016 with an ``Excellent'' degree. His main research interest is in the area of Network and network security and privacy. In this area, he published more than 100 papers in top-most international peer-reviewed journals and conference, e.g., IEEE TCC, IEEE TNSM, IEEE TGCN, and IEEE ICC/GLOBECOM (h-index=27, 2.8k+ citations). He is an Associate Editor in IEEE Transactions on Consumer Electronics, IET Communication and Cluster Computing Journals. He is a Senior Member of the IEEE. For additional information: \url{http://mshojafar.com} 
\end{minipage}

\hfill\break 

\noindent\begin{minipage}{0.14\textwidth}
\includegraphics[width=1in,height=1.15in,clip,keepaspectratio]{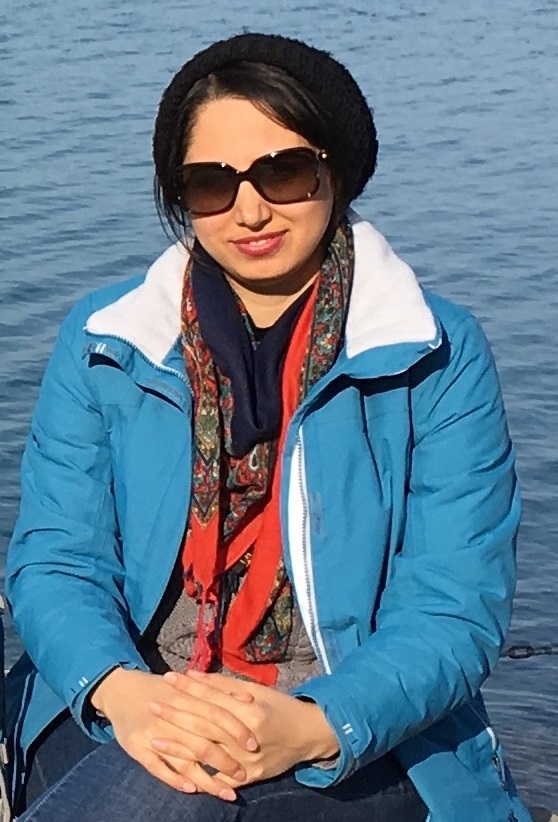}  
\end{minipage}%
\hfill
\begin{minipage}{0.84\textwidth}		
 \textbf{Zahra~Pooranian} is currently a Postdoc in the SPRITZ Security and Privacy Research group at the University of Padua, Italy, since April 2017. She received her Ph.D. degree in Computer Science Sapienza University of Rome,  Italy, in February 2017. She is a (co)author of several peer-reviewed publications (h-index=17, citations=750+) in well-known conferences and journals. She is an Editor of KSSI transaction on internet and information systems and Future Internet. Her current research focuses on Machine Learning, Smart Grid, and Cloud/Fog Computing. She was a programmer in several companies in Iran from 2009-2014, respectively. She is a member of IEEE. For additional information: \url{https://www.math.unipd.it/~zahra/} 
\end{minipage}%

\hfill \break
\clearpage

\noindent\begin{minipage}{0.14\textwidth}
\includegraphics[width=1.15in,height=1.15in,clip,keepaspectratio]{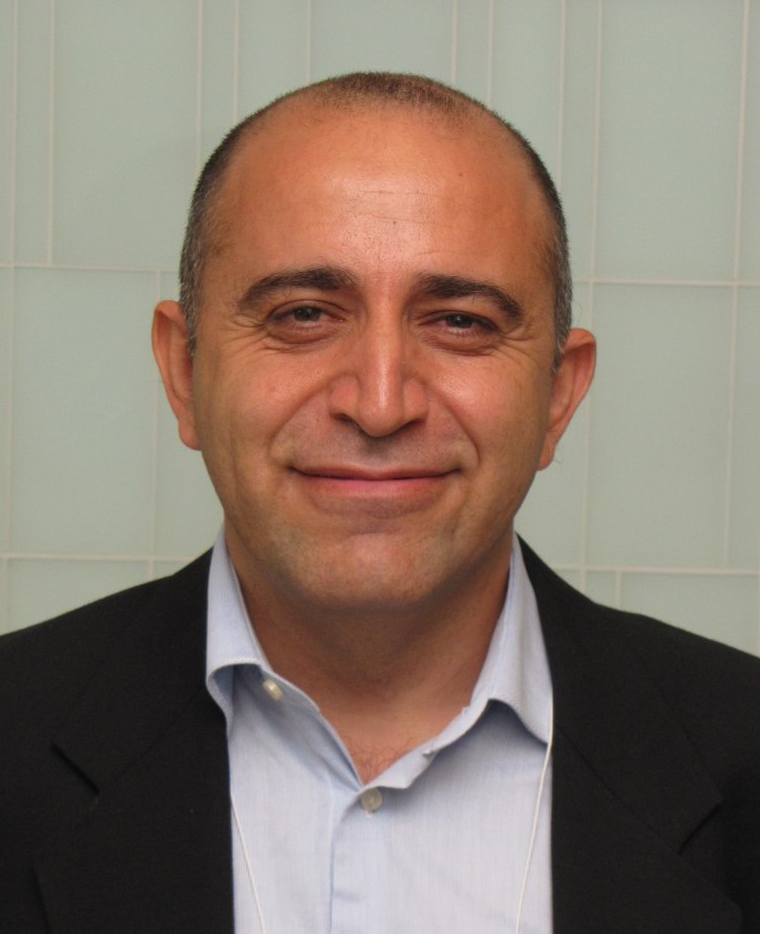}\end{minipage}%
\hfill%
\begin{minipage}{0.84\textwidth}	
\textbf{Ali Miri} has been a Full Professor at the School of Computer Science, Ryerson University, Toronto since 2009. He is the Research Director, Privacy and Big Data Institute, Ryerson University, an Affiliated Scientist at Li Ka Shing Knowledge Institute, St. Michaels Hospital, and a member of Standards Council of Canada, Big Data Working Group. He has also been with the School of Information Technology and Engineering and the Department of Mathematics and Statistics since 2001, and has held visiting positions at the Fields Institute for Research in Mathematical Sciences, Toronto in 2006, and Universite de Cergy-Pontoise, France in 2007, and Alicante and Albecete Universities in Spain in 2008. His research interests include cloud computing and big data, computer networks, digital communication, and security and privacy technologies and their applications. He has authored and co-authored more than 200 referred articles, 6 books, and 5 patents in these fields. Dr. Miri has chaired over a dozen international conference and workshops, and had served on more than 80 technical program committees. He is a senior member of the IEEE, and a member of the Professional Engineers Ontario. For additional information: \url{http://www.scs.ryerson.ca/~samiri/index.html} 
\end{minipage}

\hfill \break

\noindent\begin{minipage}{0.14\textwidth}
\includegraphics[width=1.15in,height=1.15in,clip,keepaspectratio]{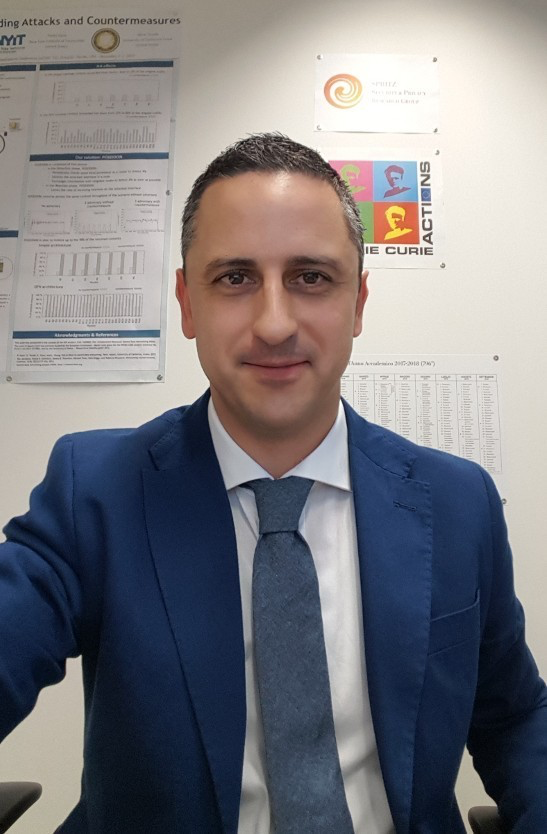}
\end{minipage}%
\hfill%
\begin{minipage}{0.84\textwidth}
\textbf{Mauro Conti} received his MSc and his PhD in Computer Science from Sapienza University of Rome, Italy, in 2005 and 2009. He has been Visiting Researcher at GMU (2008, 2016), UCLA (2010), UCI (2012, 2013, 2014), TU Darmstadt (2013), UF (2015), and FIU (2015, 2016). In 2015 he became Associate Professor, and Full Professor in 2018. He has been awarded with a Marie Curie Fellowship (2012) by the European Commission, and with a Fellowship by the German DAAD (2013). His main research interest is in the area of security and privacy. In this area, he published more than 300 papers in topmost international peer-reviewed journals and conference. He is Associate Editor for several journals, including IEEE Communications Surveys \& Tutorials, IEEE Transactions on Network and Service Management, and IEEE Transactions on Information Forensics and Security. He is Senior Member of the IEEE. For additional information: \url{http://www.math.unipd.it/~conti/}
\end{minipage}

\setcounter{figure}{0}
\setcounter{table}{0}
\renewcommand{\thefigure}{A.\arabic{figure}}
\renewcommand*\thetable{A.\arabic{table}}

\section*{Appendix}
\textcolor{black}{In this section, we explain why we select CNN classifier as the main classification algorithm applied in the paper. To this end, we tested our attack and defense algorithms against the KSSD~\cite{paudice2018label}, and GANX~\cite{taheri2019can} defense algorithms on present various classifications algorithms namely RF, SVM, DR, NN and CNN and compute accuracy and FPR metrics for different features to study on different datasets. Table~\ref{taba1} presents the results. As the results presented in this table, with the change of the classification algorithm, there is no significant difference in the superiority of one method to another. However,from this table, we conclude that the accuracy of CNN classification method for all the attack and defense algorithms compared to other classification methods in all datasets for all features is higher. Similarly, the FPR rate of CNN method is lower compared with other classification methods. As a result, \textit{we select CNN method to design our ML model for attack and defense algorithms}.}
\clearpage

\begin{table*}[!htpb]
\centering
\caption{\small \textcolor{black}{Testing the proposed methods (attacks and defenses) using training classification algorithms tested on API, Permission and Intent features in various datasets. No-A= No attack algorithm; RF= random forest; NN= neural network; SVM= support vector machine; DT= decision tree.}}
\label{taba1}
\scriptsize{
\begin{tabular}{|>{\centering\arraybackslash}m{0.1cm}|m{1.5cm}|p{0.5cm}|p{0.5cm}| p{0.5cm}| p{0.5cm}| p{0.5cm}| p{0.5cm}| p{0.5cm}| p{0.5cm}| p{0.5cm}| p{0.5cm}|}
\hline
\rowcolor{green} 
\multicolumn{12}{|c|}{\textbf{Drebin}}\\ \hline
\rowcolor{white}
&&\multicolumn{2}{|c|}{\textbf{RF}}&\multicolumn{2}{|c|}{\textbf{SVM}}&\multicolumn{2}{|c|}{\textbf{DT}}&\multicolumn{2}{|c|}{\textbf{NN}}&\multicolumn{2}{|c|}{\textbf{CNN}}\\\hline

& \textbf{Algs.}&
 \multicolumn{1}{|c|}{\textbf{Acc}}&
 \multicolumn{1}{|c|}{\textbf{FPR}}&
 \multicolumn{1}{|c|}{\textbf{Acc}}&  \multicolumn{1}{|c|}{\textbf{FPR}}&  \multicolumn{1}{|c|}{\textbf{Acc}}&  \multicolumn{1}{|c|}{\textbf{FPR}}&  \multicolumn{1}{|c|}{\textbf{Acc}}&  \multicolumn{1}{|c|}{\textbf{FPR}}&
 \multicolumn{1}{|c|}{\textbf{Acc}}&
 \multicolumn{1}{|c|}{\textbf{FPR}}\\\hline

& \cellcolor[HTML]{ffffa6}\textbf{No-A}&98.00&3.81&98.40&2.08&97.85&2.25&97.78&4.58&98.45&2.70\\\cline{2-12}

& \cellcolor[HTML]{ffffa6}\textbf{SCLFA}&83.03&28.81&83.37&27.58&82.24&27.63&82.75&29.92&83.42&28.08\\\cline{2-12}

\multirow{-1}{*}{\begin{sideways}{\textbf{API}}\end{sideways}}& \cellcolor[HTML]{ffffa6}\textbf{LSD}&90.92&17.91&91.29&16.46&90.74&16.56&90.67&18.87&91.34&17.01\\\cline{2-12}

& \cellcolor[HTML]{ffffa6}\textbf{CSD}&97.39&3.63&97.78&1.90&97.23&2.07&97.16&4.40&97.83&2.52\\\cline{2-12}

& \cellcolor[HTML]{ffffa6}\textbf{KSSD~\cite{paudice2018label}}&82.27&31.56&82.61&30.38&82.06&30.42&81.99&32.70&82.66&30.87\\\cline{2-12}

& \cellcolor[HTML]{ffffa6}\textbf{GANX~\cite{taheri2019can}}&82.82&28.31&82.90&28.38&82.54&27.72&84.67&25.79&83.30&29.07\\\hline\hline
& \cellcolor[HTML]{ffffa6}\textbf{No-A}&86.61&39.86&87.02&38.31&87.71&39.02&87.23&38.36&87.07&38.76\\\cline{2-12}

& \cellcolor[HTML]{ffffa6}\textbf{SCLFA}&74.97&59.97&75.38&58.52&76.23&58.44&75.59&58.88&75.42&58.88\\\cline{2-12}

\multirow{-3}{*}{\begin{sideways}{\textbf{Permission}}\end{sideways}}& \cellcolor[HTML]{ffffa6}\textbf{LSD}&85.85&45.10&86.26&43.59&85.90&43.04&86.47&43.71&86.30&44.01\\\cline{2-12}

& \cellcolor[HTML]{ffffa6}\textbf{CSD}&86.59&40.03&87.00&38.49&87.52&39.46&87.21&38.54&87.04&38.93\\\cline{2-12}

& \cellcolor[HTML]{ffffa6}\textbf{KSSD~\cite{paudice2018label}}&79.31&44.41&79.72&42.88&80.70&42.78&79.93&43.00&79.77&43.31\\\cline{2-12}

& \cellcolor[HTML]{ffffa6}\textbf{GANX~\cite{taheri2019can}}&79.86&45.69&80.00&40.89&83.23&43.22&82.46&39.68&80.34&43.03\\\hline\hline
& \cellcolor[HTML]{ffffa6}\textbf{No-A}&78.26&85.81&78.53&85.37&77.76&86.39&77.93&85.62&78.38&86.10\\\cline{2-12}

& \cellcolor[HTML]{ffffa6}\textbf{SCLFA}&68.75&87.04&69.01&86.68&68.25&87.54&68.41&86.87&68.86&87.30\\\cline{2-12}

\multirow{-1}{*}{\begin{sideways}{\textbf{Intents}}\end{sideways}}& \cellcolor[HTML]{ffffa6}\textbf{LSD}&75.50&87.32&75.76&86.93&74.99&87.85&75.16&87.13&75.61&87.59\\\cline{2-12}

& \cellcolor[HTML]{ffffa6}\textbf{CSD}&77.98&86.18&78.24&85.74&77.48&86.76&77.64&85.98&78.10&86.47\\\cline{2-12}

& \cellcolor[HTML]{ffffa6}\textbf{KSSD~\cite{paudice2018label}}&70.68&88.66&70.94&88.28&70.17&89.20&70.34&98.94&70.79&88.94\\\cline{2-12}

& \cellcolor[HTML]{ffffa6}\textbf{GANX~\cite{taheri2019can}}&70.53&92.07&71.22&86.13&72.56&88.16&67.96&98.90&72.82&89.21\\\hline\hline
\rowcolor{green} 
\multicolumn{12}{|c|}{\textbf{Contagio}}\\ \hline
\rowcolor{white}
&&\multicolumn{2}{|c|}{\textbf{RF}}&\multicolumn{2}{|c|}{\textbf{SVM}}&\multicolumn{2}{|c|}{\textbf{DT}}&\multicolumn{2}{|c|}{\textbf{NN}}&\multicolumn{2}{|c|}{\textbf{CNN}}\\\hline

& \textbf{Algs.}&
 \multicolumn{1}{|c|}{\textbf{Acc}}&
 \multicolumn{1}{|c|}{\textbf{FPR}}&
 \multicolumn{1}{|c|}{\textbf{Acc}}&  \multicolumn{1}{|c|}{\textbf{FPR}}&  \multicolumn{1}{|c|}{\textbf{Acc}}&  \multicolumn{1}{|c|}{\textbf{FPR}}&  \multicolumn{1}{|c|}{\textbf{Acc}}&  \multicolumn{1}{|c|}{\textbf{FPR}}&
 \multicolumn{1}{|c|}{\textbf{Acc}}&
 \multicolumn{1}{|c|}{\textbf{FPR}}\\\hline
& \cellcolor[HTML]{ffffa6}\textbf{No-A}&98.45&2.70&98.45&6.72&98.27&6.38&97.11&10.36&97.95&12.90\\\cline{2-12}

& \cellcolor[HTML]{ffffa6}\textbf{SCLFA}&75.97&9.52&75.79&9.28&74.98&11.17&75.48&15.84&75.62&12.32\\\cline{2-12}

\multirow{-1}{*}{\begin{sideways}{\textbf{API}}\end{sideways}}& \cellcolor[HTML]{ffffa6}\textbf{LSD}&89.28&7.84&89.10&7.54&88.60&8.02&88.78&14.08&88.92&10.60\\\cline{2-12}

& \cellcolor[HTML]{ffffa6}\textbf{CSD}&98.39&12.61&98.21&12.46&97.42&14.08&97.89&19.06&98.04&15.47\\\cline{2-12}

& \cellcolor[HTML]{ffffa6}\textbf{KSSD~\cite{paudice2018label}}&78.44&7.28&78.25&6.96&77.67&8.31&77.94&99.72&78.08&10.03\\\cline{2-12}

& \cellcolor[HTML]{ffffa6}\textbf{GANX~\cite{taheri2019can}}&79.11&9.27&78.60&0.86&80.49&9.46&80.72&99.67&80.43&7.84\\\hline\hline
& \cellcolor[HTML]{ffffa6}\textbf{No-A}&98.30&12.02&98.30&8.61&98.87&10.73&97.60&18.10&97.95&15.02\\\cline{2-12}

& \cellcolor[HTML]{ffffa6}\textbf{SCLFA}&72.09&65.40&72.09&62.61&72.15&68.14&71.39&73.93&71.74&69.67\\\cline{2-12}

\multirow{-3}{*}{\begin{sideways}{\textbf{Permission}}\end{sideways}}& \cellcolor[HTML]{ffffa6}\textbf{LSD}&92.83&65.40&92.83&62.61&93.29&68.14&92.13&73.93&92.48&69.67\\\cline{2-12}

& \cellcolor[HTML]{ffffa6}\textbf{CSD}&98.07&12.61&98.07&9.20&98.63&11.36&97.37&18.71&97.72&15.62\\\cline{2-12}

& \cellcolor[HTML]{ffffa6}\textbf{KSSD~\cite{paudice2018label}}&70.89&64.81&82.30&8.22&72.43&42.89&70.19&73.31&70.54&69.07\\\cline{2-12}

& \cellcolor[HTML]{ffffa6}\textbf{GANX~\cite{taheri2019can}}&70.86&74.68&82.52&7.38&75.11&43.89&73.93&56.29&70.54&68.69\\\hline\hline
& \cellcolor[HTML]{ffffa6}\textbf{No-A}&90.43&92.81&90.43&92.81&91.75&75.07&90.37&85.87&96.84&29.13\\\cline{2-12}

& \cellcolor[HTML]{ffffa6}\textbf{SCLFA}&80.49&98.11&80.49&98.11&81.80&87.50&80.51&93.75&	86.89&	61.55\\\cline{2-12}

\multirow{-1}{*}{\begin{sideways}{\textbf{Intents}}\end{sideways}}& \cellcolor[HTML]{ffffa6}\textbf{LSD}&89.06&93.24&89.06&93.24&90.43&77.00&89.01&86.90&95.47&35.77\\\cline{2-12}

& \cellcolor[HTML]{ffffa6}\textbf{CSD}&90.11&94.03&90.11&94.03&91.43&76.27&90.05&87.02&96.52&30.54\\\cline{2-12}

& \cellcolor[HTML]{ffffa6}\textbf{KSSD~\cite{paudice2018label}}&81.92&96.98&81.92&96.98&83.24&86.06&81.93&92.53&88.33&59.18\\\cline{2-12}

& \cellcolor[HTML]{ffffa6}\textbf{GANX~\cite{taheri2019can}}&82.09&90.09&82.27&92.58&81.51&89.36&84.16&75.34&91.49&49.46\\\hline

\rowcolor{green} 
\multicolumn{12}{|c|}{\textbf{Gnome}}\\ \hline
\rowcolor{white}
&&\multicolumn{2}{|c|}{\textbf{RF}}&\multicolumn{2}{|c|}{\textbf{SVM}}&\multicolumn{2}{|c|}{\textbf{DT}}&\multicolumn{2}{|c|}{\textbf{NN}}&\multicolumn{2}{|c|}{\textbf{CNN}}\\\hline

& \textbf{Algs.}&
 \multicolumn{1}{|c|}{\textbf{Acc}}&
 \multicolumn{1}{|c|}{\textbf{FPR}}&
 \multicolumn{1}{|c|}{\textbf{Acc}}&  \multicolumn{1}{|c|}{\textbf{FPR}}&  \multicolumn{1}{|c|}{\textbf{Acc}}&  \multicolumn{1}{|c|}{\textbf{FPR}}&  \multicolumn{1}{|c|}{\textbf{Acc}}&  \multicolumn{1}{|c|}{\textbf{FPR}}&
 \multicolumn{1}{|c|}{\textbf{Acc}}&
 \multicolumn{1}{|c|}{\textbf{FPR}}\\\hline
& \cellcolor[HTML]{ffffa6}\textbf{No-A}&99.37&4.08&99.16&4.96&98.59&5.04&98.89&8.02&99.52&2.94\\\cline{2-12}

& \cellcolor[HTML]{ffffa6}\textbf{SCLFA}&71.97&53.88&71.76&55.37&71.55&50.84&71.49&54.58&72.12&54.20\\\cline{2-12}

\multirow{-1}{*}{\begin{sideways}{\textbf{API}}\end{sideways}}& \cellcolor[HTML]{ffffa6}\textbf{LSD}&84.66&83.45&84.45&86.62&86.46&48.55&84.18&81.48&84.81&85.51\\\cline{2-12}

& \cellcolor[HTML]{ffffa6}\textbf{CSD}&98.83&6.94&98.62&7.85&97.03&8.02&92.76&46.67&98.98&5.88\\\cline{2-12}

& \cellcolor[HTML]{ffffa6}\textbf{KSSD~\cite{paudice2018label}}&73.18&52.77&72.97&54.31&76.16&55.70&72.70&53.57&73.33&53.07\\\cline{2-12}

& \cellcolor[HTML]{ffffa6}\textbf{GANX~\cite{taheri2019can}}&73.40&59.39&73.33&44.49&79.41&60.96&75.81&34.86&73.22&53.42\\\hline\hline
& \cellcolor[HTML]{ffffa6}\textbf{No-A}&94.72&54.69&94.93&51.63&93.91&59.24&94.54&62.66&94.57&57.14\\\cline{2-12}

& \cellcolor[HTML]{ffffa6}\textbf{SCLFA}&92.80&55.51&93.01&52.44&91.99&60.08&92.62&63.52&92.65&57.98\\\cline{2-12}

\multirow{-3}{*}{\begin{sideways}{\textbf{Permission}}\end{sideways}}& \cellcolor[HTML]{ffffa6}\textbf{LSD}&94.12&55.83&94.34&52.70&93.31&60.52&93.94&64.04&93.97&58.37\\\cline{2-12}

& \cellcolor[HTML]{ffffa6}\textbf{CSD}&94.57&55.10&94.78&52.03&93.76&59.66&94.39&63.09&94.42&57.56\\\cline{2-12}

& \cellcolor[HTML]{ffffa6}\textbf{KSSD~\cite{paudice2018label}}&92.53&55.92&92.74&52.85&91.72&60.50&92.35&63.95&92.38&58.40\\\cline{2-12}

& \cellcolor[HTML]{ffffa6}\textbf{GANX~\cite{taheri2019can}}&93.22&62.34&93.10&43.60&91.98&59.36&92.71&47.73&92.85&57.59\\\hline\hline
& \cellcolor[HTML]{ffffa6}\textbf{No-A}&93.19&84.03&92.86&80.45&93.28&83.59&93.04&82.58&99.22&8.78\\\cline{2-12}

& \cellcolor[HTML]{ffffa6}\textbf{SCLFA}&88.42&91.22&88.09&88.86&88.15&94.07&88.27&90.27&94.45&43.03\\\cline{2-12}

\multirow{-1}{*}{\begin{sideways}{\textbf{Intents}}\end{sideways}}& \cellcolor[HTML]{ffffa6}\textbf{LSD}&91.21&89.84&90.88&86.69&91.54&90.75&91.06&88.56&97.24&25.00\\\cline{2-12}

& \cellcolor[HTML]{ffffa6}\textbf{CSD}&91.33&87.82&91.00&84.76&91.53&88.20&91.18&86.58&97.36&24.44\\\cline{2-12}

& \cellcolor[HTML]{ffffa6}\textbf{KSSD~\cite{paudice2018label}}&90.55&88.86&90.22&86.05&90.13&90.60&90.40&96.07&96.58&30.88\\\cline{2-12}

& \cellcolor[HTML]{ffffa6}\textbf{GANX~\cite{taheri2019can}}&88.96&94.51&90.58&79.02&90.25&87.26&90.64&95.89&96.55&30.25\\\hline
\end{tabular}
}
\end{table*}

\end{document}